\documentclass{article}

\usepackage[preprint]{corl_2026} 

\usepackage{amsfonts}
\usepackage{amsmath}
\usepackage{graphicx}
\usepackage{multirow}
\usepackage[table]{xcolor}
\usepackage{capt-of}
\usepackage{wrapfig}
\usepackage[title]{appendix} 

\title{RoPT: Robust 6DoF Object Pose Tracking in Video}

\author{
  Yongqing Liang, Yuzhong Chen, Sihang Li\\
  New York University\\
  \texttt{yongqing.liang@nyu.edu} \\
  \And
  Junsuk Kim, Chase Kidder, Hui-ping Wang, Yunzhi Xu, Raihan Haque, Yubiao Zhang, \\
  General Motors \\
  \AND
  Chen Feng$^*$ \\
  New York University \\
  \texttt{cfeng@nyu.edu} \\
}

\begin{document}
\maketitle


\begin{abstract}
Accurate 6-DoF object pose tracking is essential for robotic interaction, yet conventional monocular pipelines frequently suffer from catastrophic divergence under rapid object motion. 
We present RoPT, a robust tracking framework engineered to overcome this limitation by unifying visual perception, temporal kinematics, and global re-localization into a tightly coupled closed loop.
Unlike existing open-loop trackers, RoPT delivers three critical capabilities that fundamentally alter system resilience. First, the visual perception component tracks object masks to robustly handle partial and severe occlusions. Second, a kinematic factor graph regularizes frame-to-frame updates, forecasts forward pose predictions, and drives a parallel multi-hypothesis verification scheme to bypass local minima. Finally, an active re-localization module transforms the pipeline into a closed-loop system, making it failure-aware and capable of deterministic self-recovery. We demonstrate the effectiveness of our framework across challenging benchmarks alongside real-world robotic demonstrations, achieving significantly higher tracking survivability under rapid motion.
\end{abstract}





\section{Introduction}
\label{sec:introduction}

\begin{wrapfigure}{r}{0.4\textwidth}
    \centering
    \vspace{-15pt}
    \includegraphics[width=\linewidth]{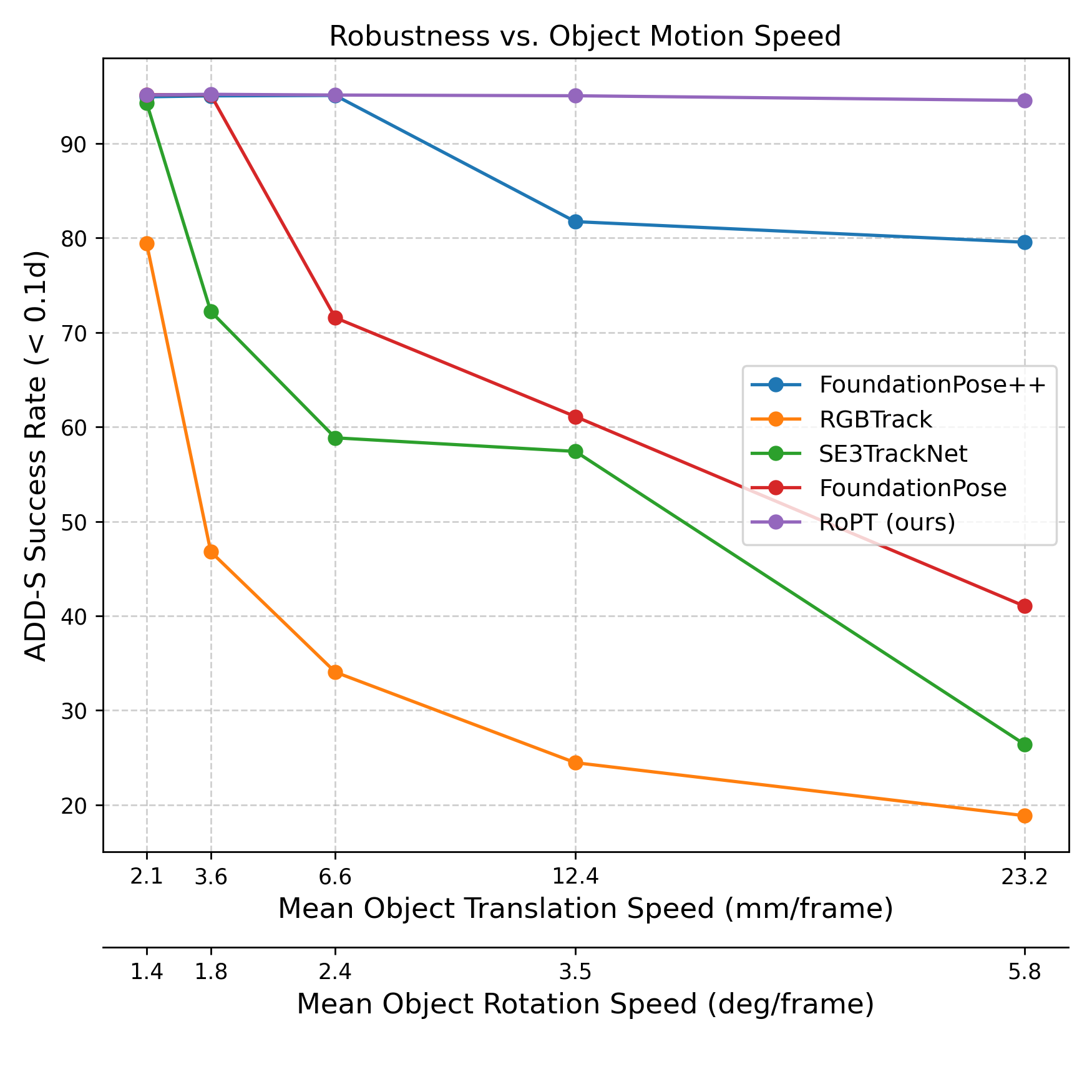}
    \vspace{-20pt}
    \caption{
    Tracking robustness evaluation on YCBInEOAT. 
    }
    \label{fig:exp-robustness-vs-motion}
    \vspace{-10pt}
\end{wrapfigure}

Robust 6DoF object pose tracking serves as a foundational pillar in modern robotics by providing the precise geometric awareness necessary for seamless human-robot collaboration and complex manipulation. 6DoF pose estimation involves determining the full rigid transformation of an object in three-dimensional space, which encompasses its translational position and rotational orientation relative to a defined coordinate frame. While recent breakthroughs in video understanding have significantly advanced 2D object localization and mask segmentation, these methods often operate purely in the image plane. Such approaches fail to provide the full spatial orientation and geometric grounding required for a robot to interact reliably with the physical world. Consequently, a critical discrepancy remains between existing video understanding frameworks and the requirements of robotic control.

The effectiveness of a tracking system depends on the two critical attributes of robustness and failure-awareness. Robustness entails the ability of an algorithm to remain accurate during high-speed motion and significant fluctuations in object appearance. Conversely, failure-awareness requires the system to actively monitor its own tracking state, dynamically quantifying uncertainty so it can flag tracking anomalies before drift compounds irrecoverably. In real-world deployments or data-driven imitation learning pipelines, losing the 6-DoF tracking of a target object is catastrophic, causing a complete breakdown in policy execution and resulting in total failure of the entire manipulation process. Despite these high stakes, conventional frame-to-frame local refinement pipelines~\citep{wen2024foundationpose, wen2020se} fail on both counts. 
When subjected to sudden high-acceleration displacements, objects frequently shear entirely out of the local tracking crop, causing visual networks to process background clutter and become trapped in false local minima. Furthermore, because these monocular trackers operate as open-loop systems, they lack the structural capacity to detect their own divergence, leading to permanent task failure. To achieve true tracking resilience under real-world dynamics, a framework must not only anticipate motion through continuous temporal regularization but also possess a deterministic mechanism to self-heal once visual tracking breaks.

To address these challenges, we introduce RoPT, a robust 6-DoF object pose tracking framework designed specifically to bridge the gap between fragile visual perception models and the rigorous demands of dynamic robotic manipulation. While techniques such as mask-based tracking, kinematic filtering, and feature-based re-localization have been explored individually in literature, isolated components are insufficient to handle the compounding errors of high-speed physical interactions. Our primary contribution to the robot learning community lies not in a single algorithmic silver bullet, but in a meticulously engineered, closed-loop system architecture that balances accuracy, resilience, and computational efficiency. We observe that robustness against rapid, erratic object motion, a ubiquitous challenge in contact-rich tasks and teleoperation,
has received disproportionately little attention due to its inherent difficulty. RoPT explicitly tackles this by tightly integrating continuous kinematic velocity forecasting with a parallel multi-hypothesis verification scheme and a deterministic, failure-triggered self-recovery module. By intelligently routing expensive computational resources only when geometric ambiguity or catastrophic tracking loss occurs, the framework achieves near real-time performance at 18.6 FPS, ensuring it remains practical for latency-sensitive control and imitation learning pipelines.
In summary, our main contributions are as follows:
\begin{enumerate}
    \item Resilience to High-Acceleration Dynamics: We propose RoPT, a closed-loop 6-DoF tracking framework that maintains unbroken trajectory consistency during rapid, erratic object motions where existing state-of-the-art trackers catastrophically diverge.
    
    \item Kinematic Factor Graph-Enabled Multi-Hypothesis: We introduce a continuous temporal graph backend that drives a parallel multi-hypothesis verification scheme, allowing the system to gracefully bypass local geometric minima and severely degraded 2D masks.
    
    \item Failure-Aware Active Self-Recovery: We design a deterministic re-localization module that automatically detects catastrophic tracking divergence and instantly self-heals by matching real-world observations against predictive synthetic CAD renderings.

\end{enumerate}

\section{Related Work}
\label{sec:related-word}

\paragraph{6-DoF Perception in Robotic Manipulation}
The integration of 6DoF object tracking has become a cornerstone for complex robotic tasks, ranging from aerial manipulation \cite{sun2022ick} to autonomous drone-based grasping \cite{sun2025towards}. Beyond basic pick-and-place, recent work has utilized high-fidelity tracking to enable visual servoing through point-tracking \cite{vecerik2024robotap} and reactive manipulation in highly dynamic environments \cite{burgess2023architecture, burgess2024reactive}. These frameworks also play a vital role in human-robot interaction via VR \cite{ajikumar20256d} and reducing uncertainty during grasping tasks \cite{bauza2024simple}. Most notably, real-time tracking has facilitated high-speed capabilities, such as catching flying objects \cite{zhang2025catch} and online replanning \cite{shao2024online}.
Despite these advancements, these applications remain tethered to the accuracy of the underlying pose estimator. Existing frameworks often struggle in real-world settings characterized by rapid motion or near-field constraints, where depth information is either unavailable in web-based video data or rendered invalid by the minimum working distance of active sensors during close-range manipulation.

\paragraph{6DoF Pose Estimation and Tracking}
Existing 6DoF perception frameworks are generally categorized into static pose estimation and temporal tracking. Static estimation is further divided into image-based methods that rely on learned categorical priors \cite{LiYangWangBasakSunGayakaSunKuo, liu2025unopose, liu2025one2any, lee2025any6d, li2025gce} and model-based methods that utilize explicit 3D geometry for alignment \cite{PengLiuHuangBaoZhou2018, MoonSonHurKim2025, wen2024foundationpose, di2021so, xu20246d, lin2024sam, wen2020se, Wenhao_Yan_and_Jie_Chu_FoundationPose_2025, guo2025rgbtrack}. While model-based approaches typically yield higher geometric precision by leveraging CAD-defined constraints, they mainly focus on local pose refinement and assume the pose changes between frames are small.
Our work moves beyond this assumption and provide a robust tracking framework under any object motions.

\paragraph{Foundation Models for Visual Perception}
Recent large-scale foundation models has revolutionized 2D perception by providing unprecedented zero-shot generalization for segmentation and feature matching, such as EdgeTAM~\cite{zhou2025edgetam}, MAST3R~\cite{leroy2024grounding} and VGGT~\cite{wang2025vggt}.
Our work bridges this gap by utilizing these foundation models as a robust perception backbone for robotic visual perception. 

\section{Method}
\label{sec:method}

\begin{figure*}[t]
    \centering
    \includegraphics[width=0.9\linewidth]{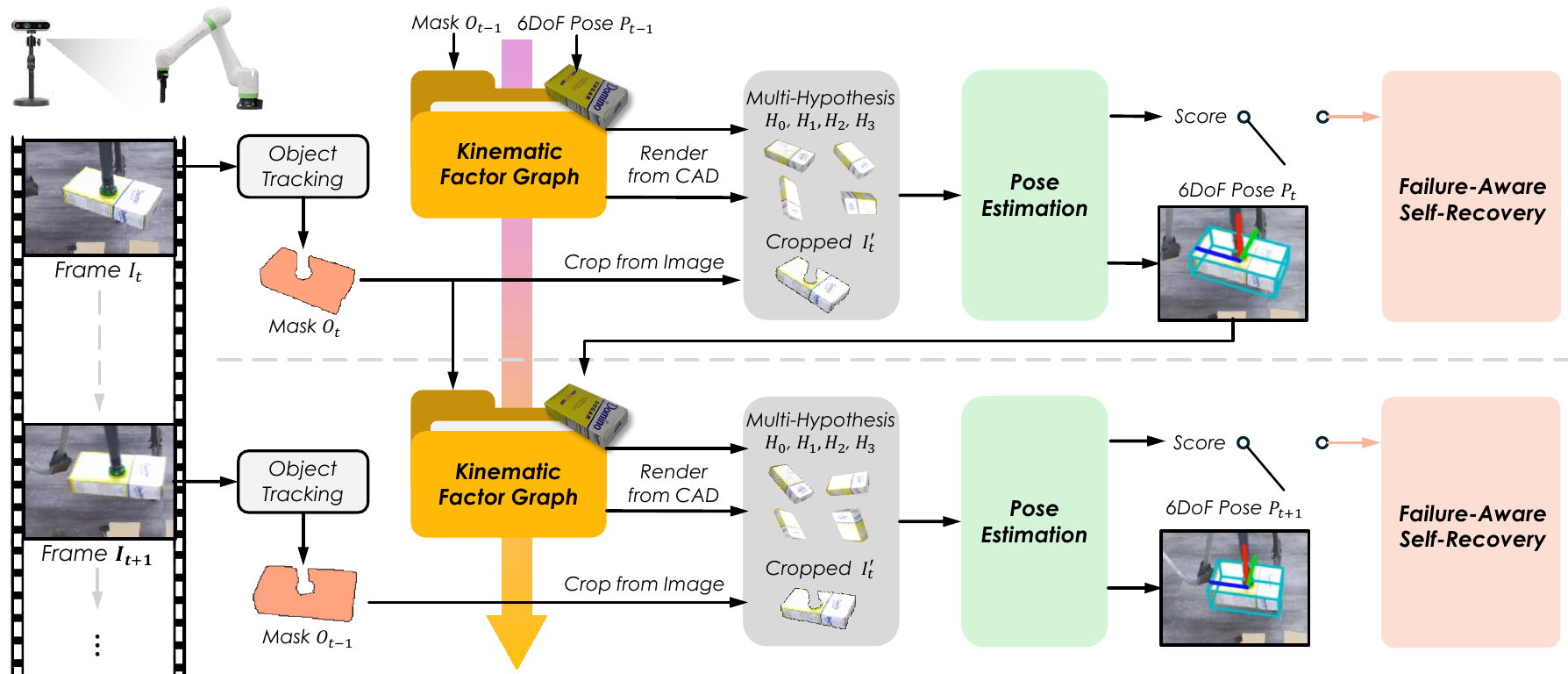}
    \caption{An Overview of the RoPT pipeline. The system integrates a mask-guided visual perception frontend with a continuous kinematic factor graph that regularizes updates and generates multi-hypothesis pose forecasts. To ensure robustness against rapid dynamics, the pipeline routes refined candidates through a pose scoring module. If catastrophic tracking loss occurs, the failure-aware self-recovery module synthesizes a virtual CAD rendering and establishes 3D-2D correspondences with the incoming RGB frame, seamlessly healing the trajectory. }
    \vspace{-0.5cm}
    \label{fig:pipeline}
    
\end{figure*}

We present RoPT, a robust framework designed for continuous 6-DoF object pose tracking in monocular RGB-D video sequences. 
Serving as a critical perception module for robotic manipulation, our pipeline explicitly prioritizes robustness against agile, high-acceleration inter-frame motions. 
This architectural focus directly addresses a fundamental vulnerability in existing trackers, which frequently suffer from catastrophic tracking loss during dynamic robot-object interactions.

Given an input RGB-D video sequence $\mathcal{V} = \{I_t\}_{t=1}^N$ and a known object 3D CAD model $\mathcal{M}$, our objective is to sequentially estimate the trajectory of poses $\mathcal{T} = \{P_t\}_{t=1}^N$. 
Each pose $P_t \in \text{SE}(3)$ represents the rigid transformation from the object-centric coordinate frame to the camera coordinate frame. 
The system must infer these transformations in a sequential, online manner while simultaneously enforcing both temporal kinematics and geometric consistency.

The RoPT framework comprises five components that function in tandem to ensure tracking stability. 
First, a baseline, render-and-compare pose refinement module iteratively aligns the CAD model to the current observation, starting from the previous frame's estimate. 
Second, a mask tracking module provides robust 2D segmentation that remains reliable under occlusion and fast motion, constraining the refinement search space. 
Third, a kinematic factor graph maintains the object's $\text{SE(3)}$ state with a constant-velocity prior, enabling multi-hypothesis generation and best-pose selection via a spread-gate scorer. 
Fourth, a global lighting optimization fits a spherical harmonics appearance model to bridge the domain gap between synthetic CAD rendering and real-world imagery. 
Fifth, a self-recovery module detects tracking failures and triggers feature-based re-localization to re-anchor the tracker before drift becomes irrecoverable. 
These five components form a closed loop to deliver robust 6-DoF tracking across challenging conditions.
Fig.~\ref{fig:pipeline} shows an overview of RoPT.

\subsection{Render-and-Compare Pose Refinement}
\label{sec:pose-refinement}

Synthesizing multi-view 2D RGB and depth images from a 3D CAD model, followed by the iterative minimization of visual and geometric alignment errors against real-world observations, represents the dominant paradigm in 6-DoF object pose estimation and tracking. 
Within this category, FoundationPose~\cite{wen2024foundationpose} establishes the current SOTA for tracking. 
While this framework exhibits high precision under slow, quasi-static object motions, its tracking accuracy degrades severely under standard or high-velocity dynamics. 
This degradation stems from a fundamental algorithmic limitation: when inter-frame displacements exceed the local basin of convergence of the refinement network, the tracker invariably converges to local minima. 

We formalize this baseline component as follows. 
Let $\mathbf{P} = \{P^{(i)}\}_{i=1}^M$ denote a set of $M$ initial pose hypotheses, and $I_t$ represent the incoming RGB-D frame at time step $t$. 
The pre-trained local refinement neural network, denoted as $\mathcal{F}_R$, iteratively predicts a relative update $\Delta P$, yielding the refined pose set:
    $\mathbf{P'} = \mathcal{F}_R(\mathbf{P}, I_t)$
Subsequently, a learned scoring network $\mathcal{F}_S$ evaluates the geometric and appearance alignment quality of each refined candidate in $\mathbf{P'}$, assigning a scalar confidence value. 
The optimal pose is selected via a winner-takes-all selection strategy:
    $P^*_t = \arg\max_{P' \in \mathbf{P'}} \mathcal{F}_S(P', I_t)$
In our subsequent ablation studies, this vanilla, open-loop frame-to-frame configuration serves as our primary baseline (denoted as \textbf{V0}).

\subsection{Occlusion-Robust Mask Tracking}
\label{sec:mask-tracking}

While the baseline (\textbf{V0}) operates effectively when an object is close to its prior pose, it scales poorly under sudden, high-velocity translation. 
In such scenarios, the object often shears entirely out of the local tracking crop, causing the refinement network to process irrelevant background clutter and drift into local minima. 
We implemented this idea that couple 2D visual tracking frontend utilizing to enforce localized spatial constraints frame-by-frame, denoted as \textbf{V1}.

In the \textbf{V1} configuration, we integrate a lightweight 2D mask tracking module to estimate a dense, binary 2D object mask from the incoming RGB stream. By combining this mask with the camera intrinsic matrix $K$, we analytically back-project the localized object depth pixels to resolve a 3D point cloud centroid, yielding an observation-driven initial translation guess that overrides the baseline prior before local $\text{SE(3)}$ refinement. Leveraging mature 2D tracking models allows our framework to robustly handle partial occlusions and temporary visual omissions, significantly expanding the tracker's tolerance to abrupt, high-velocity translations. However, as an open-loop solution, this method introduces distinct vulnerabilities: first, 2D segmentation is occasionally imperfect, where a severely degraded or broken mask can introduce corrupt translation priors that perform worse than simply relying on the previous frame's state; second, while this spatial anchoring effectively mitigates sudden translational jumps, it remains entirely blind to abrupt inter-frame rotational changes.

\subsection{SE(3) Factor Graph with Multi-Hypothesis Selection}
\label{sec:factor-graph}

While the mask-guided tracking in \textbf{V1} significantly restricts the translational search space, it remains fundamentally open-loop and susceptible to severe mask degradation. To enforce strict temporal consistency and provide predictive geometric priors, we introduce a continuous-time kinematic factor graph backend, establishing a closed-loop perception system (\textbf{V2}). We formulate the trajectory estimation as a Maximum a Posteriori (MAP) inference problem over a sliding window. The object state at time $t$ is parameterized as $P_t \in \text{SE}(3)$, with a generalized velocity $\nu_t \in \mathfrak{se}(3)$. The factor graph integrates two primary constraints:
\textbf{Kinematic Motion Factor:} We assume a constant-velocity kinematic model. To mitigate the impact of instantaneous observation noise, we track $\nu_t$ using an interval-adaptive Exponential Moving Average (EMA) on the Lie algebra manifold. The temporal constraint between consecutive frames is defined via the exponential map $P_t = P_{t-1} \exp(\nu_{t-1} \Delta t)$, yielding a physics-grounded forward prediction $P_{pred}$ at time $t$.
\textbf{Robust Observation Factor:} The pose generated by the visual frontend acts as a unary measurement prior. To prevent graph divergence from severe visual outliers, we wrap the observation covariance in a Huber robust estimator, which dynamically down-weights measurements that aggressively violate the kinematic manifold.

A critical bottleneck in standard 6-DoF refiners is their reliance on a single initial guess; if the guess falls outside the local convergence basin, or if the 2D mask is corrupted, the refinement fails. We overcome this by utilizing the factor graph to seed a parallel, multi-hypothesis refinement batch. For an incoming frame, we construct  candidate poses $\mathbf{P}_{guess} \in \mathbb{R}^{4 \times 4 \times 4}$  as follows:
\begin{itemize}
    \item $H_0$ \textit{(Static Conservative):} The optimized pose from the previous frame $(R_{prev}, t_{prev})$.
    \item $H_1$ \textit{(Dynamic Optimal):} The pure forward-simulated pose from the graph $(R_{pred}, t_{pred})$.\item $H_2$ \textit{(Predictive Anchor):} Kinematic rotation with mask-derived translation $(R_{pred}, t_{mask})$.\item $H_3$ \textit{(Static Anchor):} Previous rotation with mask-derived translation $(R_{prev}, t_{mask})$.
\end{itemize}
Following the batched refinement, the system must select the optimal pose by compute the maximum pairwise rotational divergence among the refined hypotheses or the score network $\mathcal{F}_S$ (See Supp.).

\begin{figure*}
    \centering
    \includegraphics[width=\linewidth]{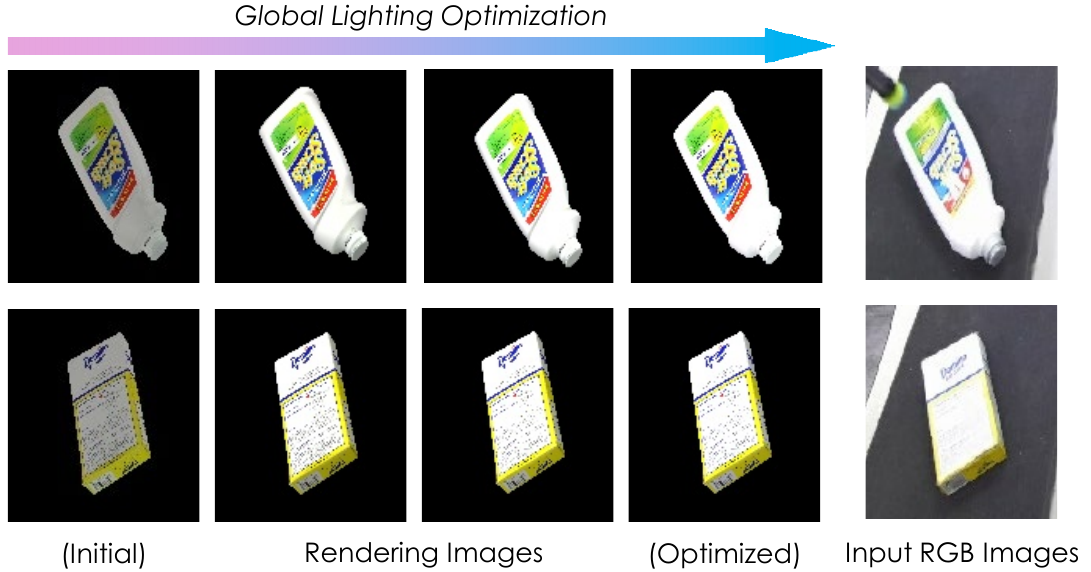}
    \caption{Visualization of the global lighting optimization. It adapts the static rendering to match real-world environmental lighting.}
    \label{fig:tex-updating}
\end{figure*}

\subsection{Global Lighting Optimization}
\label{sec:appearance}

To bridge the persistent domain gap between idealized synthetic CAD renderings and real-world imagery, a discrepancy primarily driven by unmodeled environmental illumination and material properties, we incorporate a global appearance optimization module. 
We denote this variant as \textbf{V3}.
We approximate the scene's diffuse lighting using a view-independent, second-order Spherical Harmonics (SH) irradiance model parameterized by 27 learnable coefficients (9 basis functions per RGB channel). 
During the initialization of the first frame, we freeze the object's known 6-DoF pose and optimize these SH coefficients during test time.
we minimize a composite objective combining a robust $L_1$ photometric residual within the visible object silhouette $\Omega$ and a band-progressive $L_2$ penalty:
\begin{equation}
\mathcal{L}(\mathbf{c}) = \frac{1}{|\Omega|} \sum_{\mathbf{u} \in \Omega} \left\| \hat{I}(\mathbf{u}, \mathbf{c}, P_{init}) - I_{obs}(\mathbf{u}) \right\|_1 + \lambda_1 \|\mathbf{c}^{(1)}\|_2^2 + \lambda_2 \|\mathbf{c}^{(2)}\|_2^2, 
\label{eqn:lighting}
\end{equation}
where $\hat{I}$ is the differentiable CAD rendering under the anchored initial pose $P_{init}$, and $\mathbf{c}^{(1)}, \mathbf{c}^{(2)}$ denote the first- and second-order SH bands.
Once converged, these global lighting parameters are locked and applied to all renderings throughout the remainder of the video pose tracking. 

\begin{wrapfigure}{r}{0.5\textwidth}
    \vspace{-20pt}
    \centering
    \includegraphics[width=\linewidth]{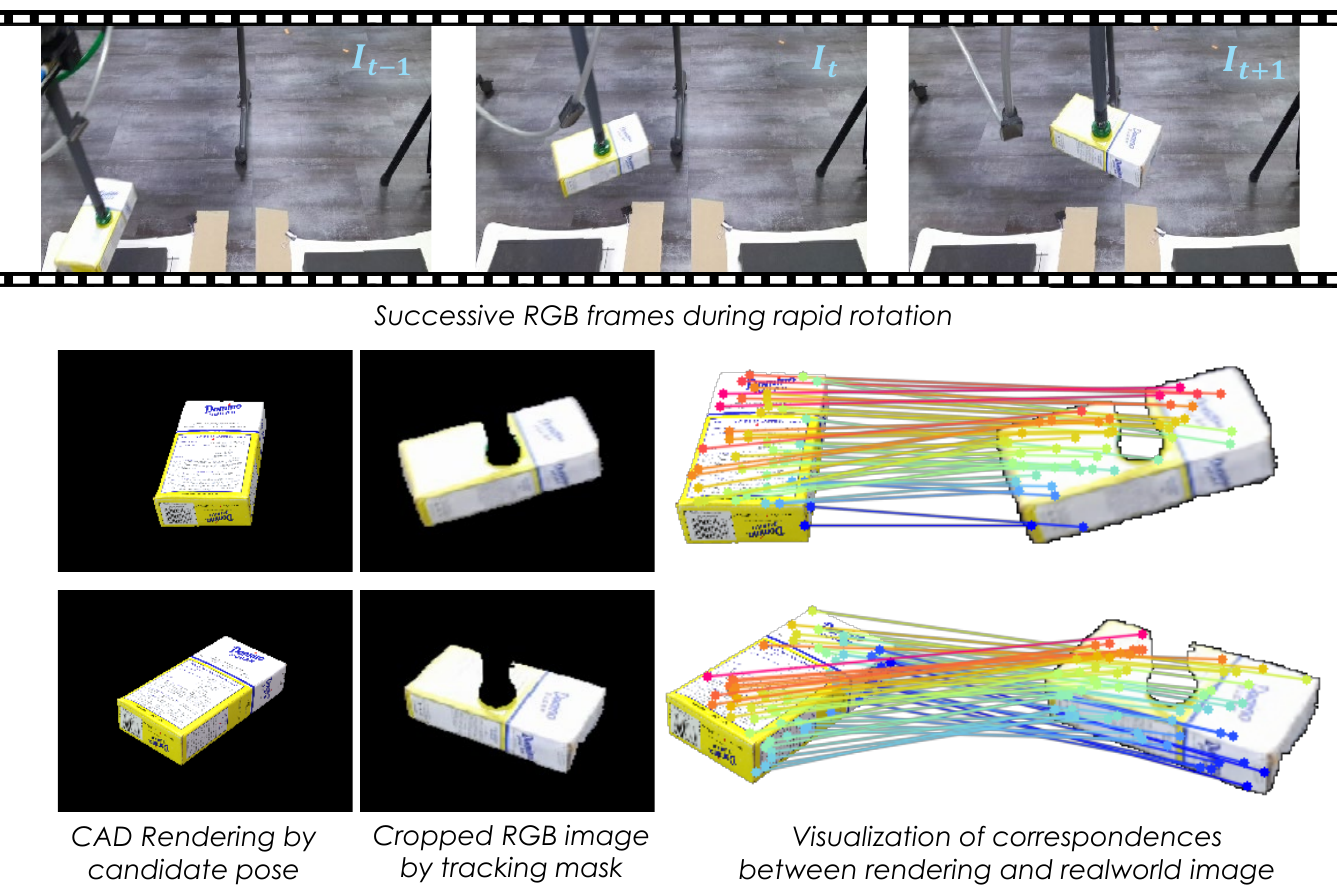}
    \caption{Pose recovery via feature matching. 
    }
    \label{fig:recovery-matching}
    \vspace{-10pt}
\end{wrapfigure}

\subsection{Failure-Aware Self-Recovery}
Despite the robust continuous tracking provided by the factor graph backend, extreme inter-frame motion or severe object occlusion can still cause irreversible divergence. To ensure long-term autonomy in embodied manipulation tasks, we introduce an active re-localization module, establishing our final model architecture (denoted as \textbf{V4}). This recovery mechanism is actively triggered when the multi-hypothesis alignment score from the scoring network $\mathcal{F}_S$ exceeds a critical threshold, a metric strongly correlated with rapid tracking failure.
Once triggered, our recovery module utilizes the current predictive pose 
to synthesize a virtual CAD rendering. 
We then deploy a dense feature matcher (Mast3R) to establish correspondences between this predictive synthetic render and the incoming real-world observation query. 
Because the rendered color buffer and the underlying 3D object-frame coordinates are generated simultaneously in a single rasterization pass, they share perfect pixel alignment. This allows us to instantly lift the 2D-to-2D dense matches into 3D-to-2D correspondences. We use standard Perspective-n-Point (PnP) with RANSAC to recover 6-DoF Pose, and update back into the factor graph to re-anchor the perception system.

\section{Experiments}
\label{sec:experiments}

\begin{table*}[t]
    \centering
    \resizebox{0.85\textwidth}{!}{%
    \begin{tabular}{l|cc|cc|cc|cc|cc|ccc}
    
    \multicolumn{13}{l}{Mean 2.1 mm translation and $1.4^\circ$ rotation per frame (Original Framerate).}\\
    \hline
        \multirow{2}{*}{Approach} &  \multicolumn{2}{c|}{cracker box} & \multicolumn{2}{c|}{bleach cleanser} & \multicolumn{2}{c|}{sugar box}  & \multicolumn{2}{c|}{tomato soup can} & \multicolumn{2}{c|}{mustard bottle} & \multicolumn{2}{c}{All}\\
        & ADD & ADD-S  & ADD & ADD-S  & ADD & ADD-S  & ADD & ADD-S  & ADD & ADD-S  & ADD & ADD-S \\
    \hline
        TrackNet & 87.36 & 90.92 & 92.19 & 95.73 & 89.75 & 92.82 & 92.12 & 95.57 & 98.63 & 98.76 & 91.34 & 94.28\\
        RGBTrack & 1.38 & 5.98 & 29.68 & 54.06 & 66.99 & 81.17 & 71.53 & 85.54 & 80.15 & 89.42 & 64.41 & 79.40\\
        FPose & 88.36 & 92.25 & 93.17 & 96.35 & 95.01 & 97.19 & 90.65 & 95.04 & 96.14 & 97.16 &  \cellcolor{gray!30}91.96 & \cellcolor{cyan!15}\textbf{95.16} \\
        FPose++ & 88.26 & 91.95 & 92.93 & 96.02 & 94.09 & 96.69 & 90.47 & 94.99 & 96.24 & 97.22 & 91.70 & 94.94 \\
        \textbf{RoPT} & 88.37 & 92.19 & 93.15 & 96.13 & 95.13 & 97.21 & 90.69 & 95.08 & 96.28 & 97.24 & \cellcolor{cyan!15}\textbf{92.02} & \cellcolor{gray!30}95.14 \\
    \hline

    \\
    \multicolumn{13}{l}{Mean 3.6 mm translation and $1.8^\circ$ rotation per frame ($2\times$ Framerate Downsampling).}\\
    \hline
        \multirow{2}{*}{Approach} &  \multicolumn{2}{c|}{cracker box} & \multicolumn{2}{c|}{bleach cleanser} & \multicolumn{2}{c|}{sugar box}  & \multicolumn{2}{c|}{tomato soup can} & \multicolumn{2}{c|}{mustard bottle} & \multicolumn{2}{c}{All}\\
        & ADD & ADD-S  & ADD & ADD-S  & ADD & ADD-S  & ADD & ADD-S  & ADD & ADD-S  & ADD  & ADD-S \\
    \hline
        TrackNet &  86.06 & 89.40 & 63.51 & 66.95 & 22.83 & 23.51 & 92.14 & 95.57 & 11.86 & 65.17 & 61.84  & 72.22 \\
        RGBTrack & 0.42 & 0.81 & 33.26 & 58.46 & 8.89 & 15.81 & 68.12 & 82.78 & 8.64 & 11.39 & 34.57 & 46.79 \\ 
        FPose & 88.46 & 92.48 & 93.36 & 96.44 & 95.03 & 97.19 & 90.69 & 95.05 & 96.12 & 97.15 & \cellcolor{gray!30}92.00 & \cellcolor{cyan!15}\textbf{95.23 }  \\
        FPose++ & 88.20 & 91.93 & 92.87 & 96.05 & 95.06 & 97.20 & 90.55 & 95.02 & 96.26 & 97.23 & 91.88 & 95.04 \\
        \textbf{RoPT} & 88.39 & 92.32 & 93.35 & 96.35 & 95.21 & 97.24 & 90.72 & 95.09 & 96.31 & 97.26 & \cellcolor{cyan!15}\textbf{92.05} & \cellcolor{gray!30}95.20  \\
    \hline

    \\
    \multicolumn{13}{l}{Mean 6.6 mm translation and $2.4^\circ$ rotation per frame ($4\times$ Framerate Downsampling).}\\
    \hline
        \multirow{2}{*}{Approach} &  \multicolumn{2}{c|}{cracker box} & \multicolumn{2}{c|}{bleach cleanser} & \multicolumn{2}{c|}{sugar box}  & \multicolumn{2}{c|}{tomato soup can} & \multicolumn{2}{c|}{mustard bottle} & \multicolumn{2}{c}{All}\\
        & ADD & ADD-S  & ADD & ADD-S  & ADD & ADD-S  & ADD & ADD-S  & ADD & ADD-S  & ADD  & ADD-S \\
    \hline
        TrackNet & 85.49 & 88.87 & 23.70 & 30.01 & 22.78 & 22.94 & 92.10 & 95.53 & 10.69 & 11.13 & 56.09 & 58.84 \\
        RGBTrack & 0.00 & 0.00 & 25.36 & 46.80 & 10.64 & 16.84 & 40.23 & 53.77 & 8.34 & 10.39 & 23.28 & 34.04\\
        FPose & 86.93 & 90.44 & 91.98 & 96.08 & 44.65 & 61.30 & 90.73 & 95.06 & 11.17 & 12.23 & 69.38 & 75.26\\
        FPose++ & 88.32 & 92.03 & 92.95 & 96.39 & 95.12 & 97.23 & 90.43 & 94.99 & 96.26 & 97.23 &  \cellcolor{cyan!15}\textbf{91.89} & \cellcolor{gray!30}95.08\\
        \textbf{RoPT} & 88.30 & 92.76 & 93.04 & 96.46 & 93.28 & 96.39 & 90.75 & 95.10 & 96.24 & 97.23 & \cellcolor{gray!30}91.58 & \cellcolor{cyan!15}\textbf{95.13} \\
    \hline

    \\
    \multicolumn{13}{l}{Mean 12.4 mm translation and $3.5^\circ$ rotation per frame ($8\times$ Framerate Downsampling).}\\
    \hline
        \multirow{2}{*}{Approach} &  \multicolumn{2}{c|}{cracker box} & \multicolumn{2}{c|}{bleach cleanser} & \multicolumn{2}{c|}{sugar box}  & \multicolumn{2}{c|}{tomato soup can} & \multicolumn{2}{c|}{mustard bottle} & \multicolumn{2}{c}{All}\\
        & ADD & ADD-S  & ADD & ADD-S  & ADD & ADD-S  & ADD & ADD-S  & ADD & ADD-S  & ADD  & ADD-S \\
    \hline
        TrackNet & 84.23 & 88.18 & 22.64 & 27.66 & 23.25 & 23.34 & 69.14 & 92.06 & 10.64 & 11.70 & 49.49 & 57.41 \\
        RGBTrack & 0.00 & 0.00 & 21.75 & 44.33 & 13.28 & 18.11 & 16.52 & 27.81 & 9.70 & 10.72 & 14.68 & 24.46 \\
        FPose &  86.08 & 89.47 & 64.96 & 67.44 & 23.45 & 23.77 & 74.97 & 85.03 & 11.23 & 11.42 & 57.23 & 61.40  \\
        FPose++ & 86.46 & 90.41 & 93.59 & 96.41 & 95.22 & 97.28 & 90.54 & 95.04 & 10.25 & 11.51 & \cellcolor{gray!30}78.59 & \cellcolor{gray!30}81.73 \\
        \textbf{RoPT} & 88.65 & 92.78 & 93.30 & 96.36 & 93.56 & 96.11 & 90.76 & 95.10 & 96.32 & 97.27 & \cellcolor{cyan!15}\textbf{91.64} & \cellcolor{cyan!15}\textbf{95.05} \\
    \hline

    \\
    \multicolumn{13}{l}{Mean 23.2 mm translation and $5.8^\circ$ rotation per frame ($16\times$ Framerate Downsampling).}\\
    \hline
        \multirow{2}{*}{Approach} &  \multicolumn{2}{c|}{cracker box} & \multicolumn{2}{c|}{bleach cleanser} & \multicolumn{2}{c|}{sugar box}  & \multicolumn{2}{c|}{tomato soup can} & \multicolumn{2}{c|}{mustard bottle} & \multicolumn{2}{c}{All}\\
        & ADD & ADD-S  & ADD & ADD-S  & ADD & ADD-S  & ADD & ADD-S  & ADD & ADD-S  & ADD  & ADD-S  \\
    \hline
        TrackNet & 30.67 & 38.34 & 22.66 & 26.69 & 24.12 & 24.20 & 25.20 & 27.46 & 10.54 & 12.65 & 23.39 & 26.40 \\
        RGBTrack & 0.00 & 0.00 & 17.02 & 35.83 & 17.14 & 21.37 & 9.15 & 15.59 & 9.65 & 11.87 & 12.05 & 18.85  \\
        FPose & 86.04 & 89.48 & 26.38 & 37.97 & 24.94 & 25.27 & 24.96 & 26.59 & 12.16 & 12.35 & 38.87 & 42.00 \\
        FPose++ & 86.13 & 89.44 & 90.75 & 95.79 & 38.53 & 89.19 & 90.58 & 95.08 & 10.18 & 10.32 & \cellcolor{gray!30}67.37 & \cellcolor{gray!30}79.55\\
        \textbf{RoPT} & 87.91 & 92.65 & 93.13 & 96.08 & 90.77 & 94.44 & 90.78 & 95.09 & 96.34 & 97.29 & \cellcolor{cyan!15}\textbf{90.85} & \cellcolor{cyan!15}\textbf{94.55} \\
    \hline
    \end{tabular}
    }
    \caption{Quantitative robustness and accuracy comparison on the YCBInEOAT benchmark. }
    \label{tab:ycbineoat-cmp}
\end{table*}

\begin{table*}[t]
    \centering
    \resizebox{0.9\textwidth}{!}{%
    \begin{tabular}{l|c|c|c|c|c}
    \hline
    Method & TrackNet & RGBTrack & FoundationPose & FoundationPose++ & \textbf{RoPT}\\
    \hline
    Runtime & 30.1 ms (33.2 fps) & 37.1 ms (27.0 fps) & 23.3 ms (42.9 fps) & 35.7 ms (28.0fps) & 53.7 ms (18.6 fps)\\
    \hline
    \end{tabular}
    }
    \caption{Quantitative tracking runtime comparison on the YCBInEOAT benchmark. }
    \label{tab:runtime}
\end{table*}

We evaluated our framework on the public pose tracking benchmarks YCBInEOAT~\cite{wen2020se} and YCB-Video~\cite{xiang2017posecnn} (See Supp.), which features scenarios of human-robot and dual-robot interaction. 
Because the original sequences often exhibit slower motion than typical human-robot collaboration, we synthesized normal to high-velocity sequences by downsampling the video frames by factors of $2\times$ up to $16\times$. We reported the average motion speedd in the top of each sub-table in Tab.~\ref{tab:ycbineoat-cmp}, which mimics the normal and rapid movements encountered in real-world environments. 
We employ the standard ADD and ADD-S metrics~\cite{wen2024foundationpose} to measure 6DoF pose accuracy. These metrics quantify the geometric alignment between the estimated and ground truth poses, with ADD-S specifically accounting for object symmetry. Area Under the Curve (AUC) is reported for both metrics to provide a comprehensive assessment of tracking precision.

Our framework utilizes EdgeTAM~\cite{zhou2025edgetam} for mask tracking (\textbf{V1} to \textbf{V4}), GTSAM~\cite{gtsam, factor_graphs_for_robot_perception} for kinematics factor graph (\textbf{V2} to \textbf{V4}), Spherical Harmonic for global lighting (\textbf{V3} to \textbf{V4}), and Mast3R~\cite{leroy2024grounding} for failure recovery (\textbf{V4}). 
We used CAD models provided by the dataset, which are a ubiquitous standard in industry and manufacturing. 
During evaluation, our model doesn't see or train on these CAD model before and conducted zero-shot tracking without finetuning. 
Under purely CAD-free, recent reconstruction tools (i.e. BundleSDF~\cite{wen2023bundlesdf} and TRELLIS~\cite{xiang2025trellis2}) can provide the 3D geometry as preprocessing.
More implementation details are in Supp.

\begin{table*}[t]
    \centering
    \resizebox{\textwidth}{!}{%
    \begin{tabular}{l|cc|cc|cc|cc|cc|ccc}
    
    \hline
        \multirow{2}{*}{Setting} &  \multicolumn{2}{c|}{Original Framerate} & \multicolumn{2}{c|}{$2\times$ Downsampling} & \multicolumn{2}{c|}{$4\times$ Downsampling}  & \multicolumn{2}{c|}{$8\times$ Downsampling} & \multicolumn{2}{c|}{$16\times$ Downsampling} & \multicolumn{3}{c}{Mean}\\
         & ADD & ADD-S  & ADD & ADD-S  & ADD & ADD-S  & ADD & ADD-S  & ADD & ADD-S & ADD & ADD-S & Runtime\\
    \hline
        V0: FoundationPose (baseline) & 91.97 & 95.16 & 91.94 & 95.16 & 65.81 & 71.56 & 56.74 & 61.10 & 37.97 & 41.02 & 68.89 & 72.80 & 23.3 ms \\
        V1: V0 + Tracking Mask-Guided & 69.91 & 89.85 & 70.18 & 89.29 & 71.45 & 89.87 & 77.50 & 89.39 & 72.21 & 89.95 & 72.25 & 89.67 & 38.6 ms\\
        V2: V1 + Kinematic Factor Graph & 91.96 & 95.15 & 91.95 & 95.17 & 91.35 & 95.06 & 91.68 & 95.15 & 81.76 & 91.06 & 89.74 & 94.32 & 44.1 ms \\
        V3: V2 + Global Lighting Opt. & 92.03 & 95.17 & 92.05 & 95.20 & 91.56 & 95.13 & 91.78 & 95.22 & 81.68 & 91.13 & \cellcolor{gray!30}89.82 & \cellcolor{gray!30}94.37 & 48.1 ms\\
        \textbf{V4: V3 + Self-Recovery (ours)} & 92.02 & 95.14 & 92.05 & 95.20 & 91.58 & 95.13 & 91.64 & 95.05 & 90.85 & 94.55 & \cellcolor{cyan!15}\textbf{91.63} & \cellcolor{cyan!15}\textbf{95.02} & 53.7 ms\\
    \hline
    \end{tabular}
    }
    \caption{Ablation on the contributions of individual component on the YCBInEOAT benchmark.}
    \label{tab:ablation}
\end{table*}

\subsection{Benchmark Comparison}

\begin{figure}[t]
    \centering
    \includegraphics[width=\linewidth]{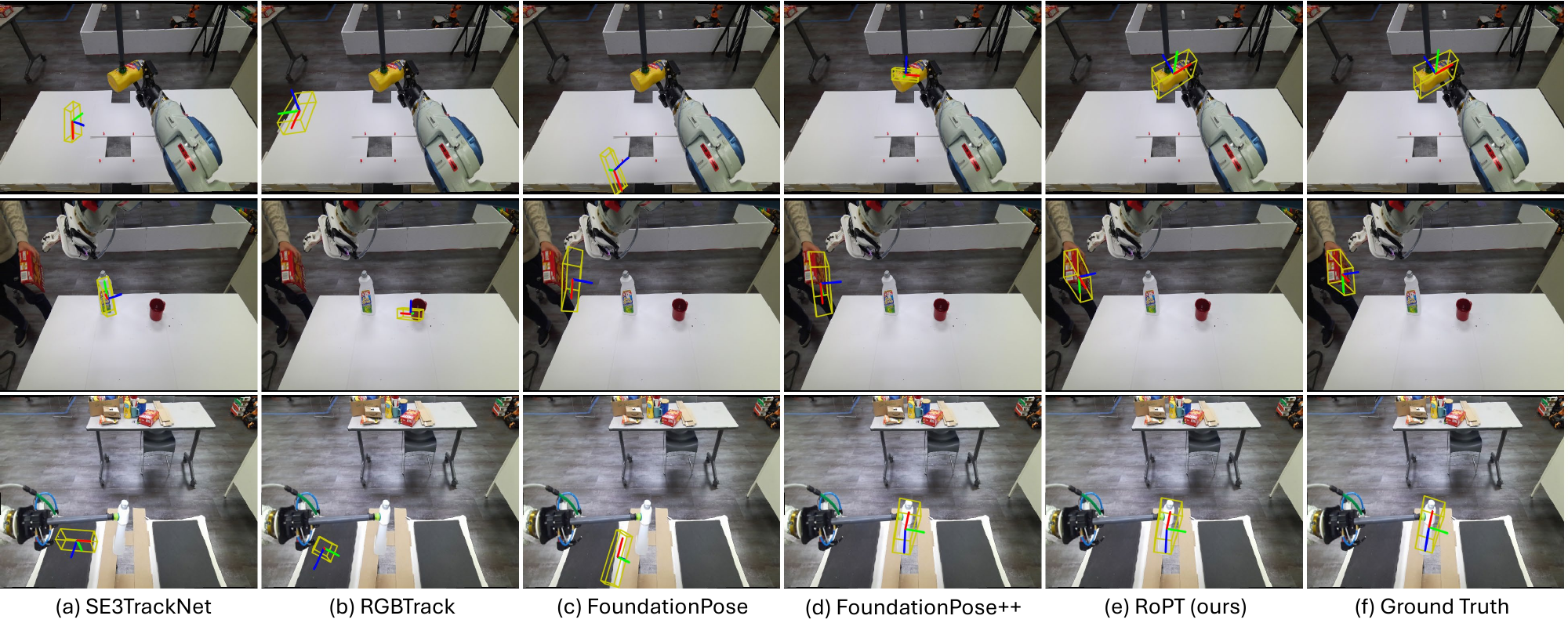}
    \caption{Qualitative tracking comparison with state-of-the-art papers on the YCBInEOAT dataset.}
    \label{fig:exp-cmp}
\end{figure}

We evaluate the performance of RoPT against open-source state-of-the-art 6DoF video trackers on the enhanced YCBInEOAT benchmark. 
We selected the model-based RGB-D methods TrackNet~\cite{wen2020se}, FoundationPose (FPose)~\cite{wen2024foundationpose}, RGBTrack~\cite{guo2025rgbtrack}, and open-source code repo FPose++~\cite{Wenhao_Yan_and_Jie_Chu_FoundationPose_2025}.
As illustrated in Fig.~\ref{fig:exp-robustness-vs-motion}, our framework exhibits a unique stability profile regardless of the motion velocity, where other methods suffer from rapid divergence.
We report the detailed comparison in Tab.~\ref{tab:ycbineoat-cmp}.
Fig.~\ref{fig:exp-cmp} shows four examples when SOTA methods failed to track the object pose. Note that once the object pose is lost, it is almost impossible to recover it.
Our method achieves state-of-the-art tracking accuracy under both slow and fast motion.
The benchmark results of YCB-Video and a new realworld dataset are in Supp.

\subsection{Ablation Study}
To validate the architectural design of RoPT, we conduct a comprehensive ablation study incrementally evaluating the framework from the open-loop baseline (\textbf{v0}) to our fully integrated closed-loop system (\textbf{v4}). 
As emphasized in our introduction, our primary contribution is not a single algorithmic anomaly, but rather a cohesive, highly robust system where each meticulously engineered component solidifies the pipeline against catastrophic failure. As shown in Tab.~\ref{tab:ablation}, the most substantial leaps in tracking survivability and overall accuracy are driven by the mask-guided tracking frontend~(\textbf{v1}), the kinematic factor graph backend~(\textbf{v2}), and the active self-recovery module~(\textbf{v4}). 
These three components form the critical backbone required to withstand rapid, high-acceleration dynamics. 
In contrast, the global lighting optimization (\textbf{v3}) yields only marginal quantitative improvements. We attribute this plateau to the baseline neural refinement module, which has already acquired strong invariance to illumination shifts through extensive data augmentation during its training phase. Nevertheless, this lighting optimization remains a structurally solid component, explicitly serving to minimize the photometric domain gap required for the downstream synthetic-to-real feature matching in the self-recovery module.

\begin{figure*}
    \centering
    \begin{tabular}{cc}
        \includegraphics[width=0.7\linewidth]{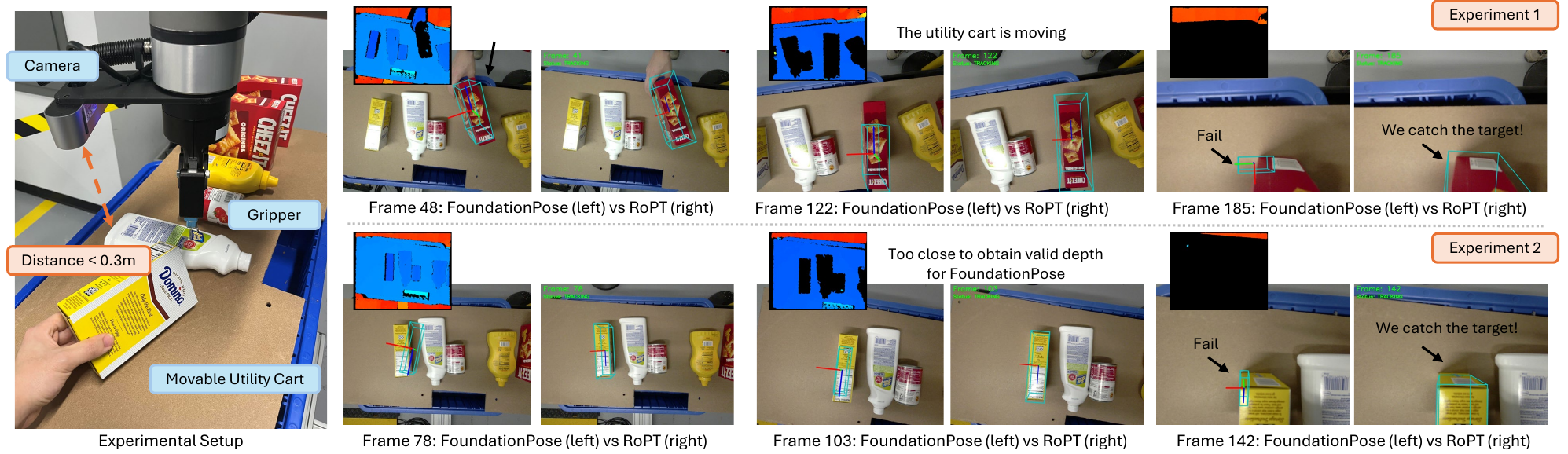} &  
        \includegraphics[width=0.23\linewidth]{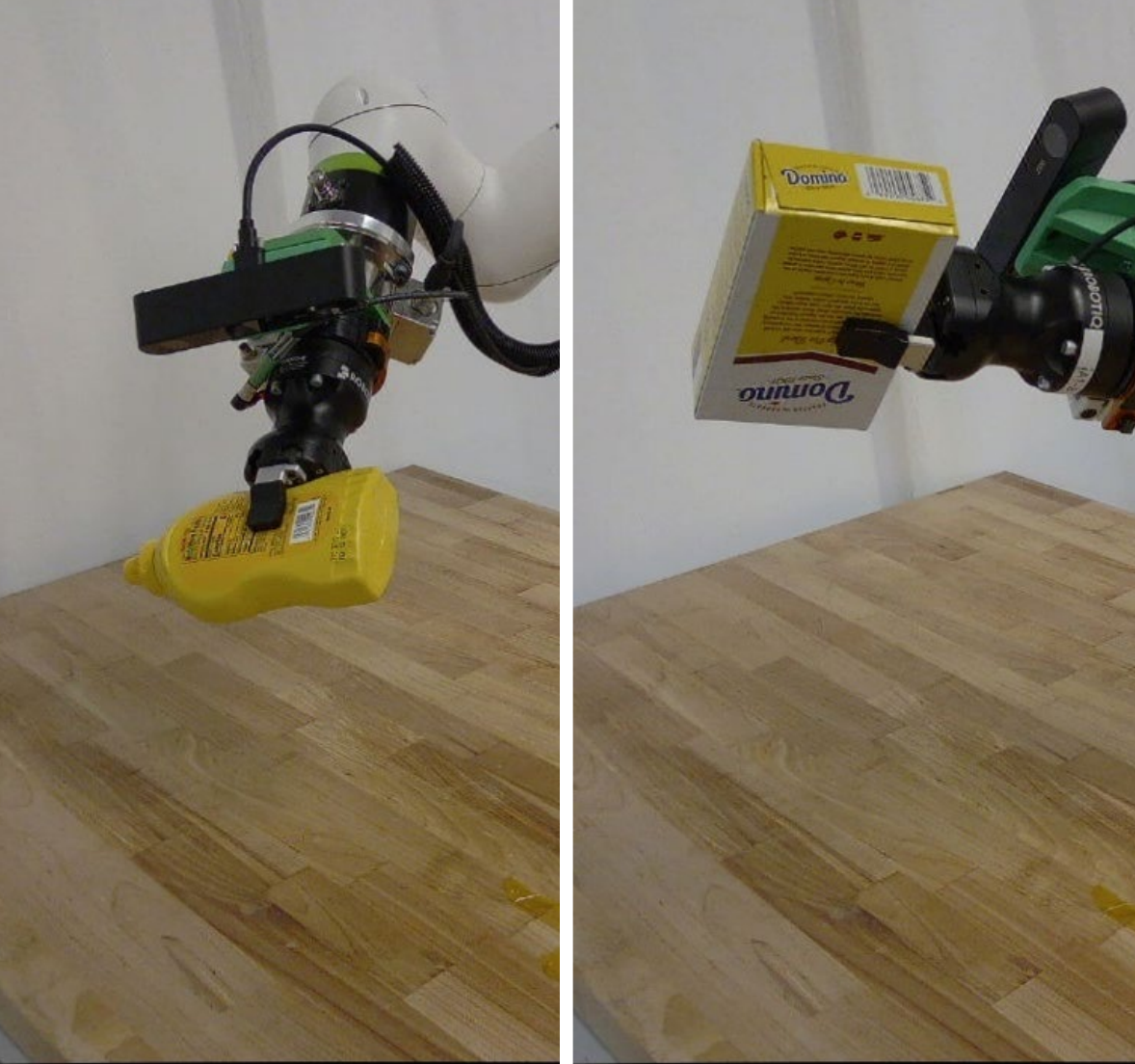} \\
         (a) Robotic tracking under dynamic motion. & (b) Tracking benchmark.
    \end{tabular}
    
    \caption{Two real-world pose tracking experiments: Grasping and Tracking Benchmark (Supp.). 
    }
    \label{fig:exp-real}
\end{figure*}

\subsection{Real-World Robotic Experiments}
We deployed RoPT on a robotic manipulator equipped with a wrist-mounted Astra 2 camera~\cite{orbbecastra2}. 
The framework provides online 6-DoF estimation to guide the gripper toward the target. For comparison, we recorded synchronized RGB-D data to evaluate the performance of FPose.
As shown in Fig.~\ref{fig:exp-real}(a), target objects were placed on a utility cart subjected to random displacements, while a human operator simultaneously applied erratic translations and rotations. This setup evaluates the system's resilience under the compounded motion of both the sensor and the target.
Our experiments highlight critical advantage of the proposed framework in robotic manipulation:
(1) Motion Robustness: While FPose loses the target due to the lack of predictive modeling during rapid displacement, RoPT maintains a persistent lock by anticipating the object’s kinematic state.
(2) Sensing Versatility: As the gripper approaches the target, the depth sensor exceeds its minimum working distance, causing near-field blindness and invalid depth reading. RoPT trigger the failure recovery to estimate the 6-DoF pose by RGB matching, ensuring precise and reliable tracking throughout the final stages of the manipulation sequence.
Furthermore, we developed a 6-DoF pose tracking benchmark categorized into three distinct velocity profiles to rigorously evaluate tracking resilience (see Supp.).
\section{Conclusion}
\label{sec:conclusion}
In this work, we presented RoPT, a resilient, closed-loop 6-DoF object pose tracking framework engineered to withstand rapid, high-acceleration dynamics in embodied systems. By tightly integrating visual perception with a continuous kinematic factor graph and a failure-aware active self-recovery module, RoPT systematically bridges the gap between fragile open-loop monocular pipelines and the rigorous demands of real-world robotic manipulation.

\paragraph{Limitations.}
While RoPT demonstrates strong resilience under dynamic conditions, its architecture relies on several limiting assumptions and exhibits specific failure modes. The system fundamentally assumes the availability of an exact geometric CAD prior and relies on RGB-D sensing, limiting its applicability on highly deformable objects, transparent materials, or specular surfaces. In terms of operational failure modes, if an object experiences severe, prolonged occlusion that outlasts the predictive horizon of the continuous kinematic factor graph, the target may drift beyond the search radius of the multi-hypothesis spread-gate, preventing autonomous self-recovery. Additionally, a practical system limitation lies in computational scalability: while the framework operates reliably at 18.6 FPS, further acceleration is currently bottlenecked by the complexity of tuning the factor graph's optimization parameters to perfectly balance temporal resilience with computational overhead. Future work will explore learning-based graph tuning and feature distillation to relax these exact-model dependencies, further advancing reliable perception for robotic intelligence.

\clearpage


\bibliography{references}  

\clearpage
\begin{appendices}
\section{Implementation Details}
We conducted experiments on NVIDIA 6000 Ada and Ubuntu 22.04 systems. 
The source code was implemented in PyTorch, and will be released to the robotic community.
Our RoPT used the full (\textbf{V4}) configuration to compare with state-of-the-art methods.
Our pipeline utilized the pretrained weights of pose refinement $\mathcal{F}_R$ and score $\mathcal{F}_S$ modules (from FoundationPose~\cite{wen2024foundationpose}, mask tracking (from EdgeTAM~\cite{zhou2025edgetam}), and feature matching (from Mast3r~\cite{leroy2024grounding}). 
It already reached the new state-of-the-art performance. 
Although further end-to-end finetuning may boost the accuracy, it is out of the scope of current paper.

We used the default settings of EdgeTAM for mask tracking and used the \emph{ISAM2} and \emph{NonlinearFactorGraph} of GTSAM as the Kinematic Factor Graph. We maintain the pose in $\text{SE(3)}$ space and the initial observation sigmas are set to $\{0.02, 0.02, 0.02, 0.10, 0.10, 0.10\}$, where the motion noise are set to all $0.05$.
Each time we have an estimated pose, we add it to the graph and optimize the graph by the API of \emph{ISAM}.

\begin{figure}[h]
    \centering
    \begin{tabular}{c}
         \includegraphics[width=0.75\linewidth]{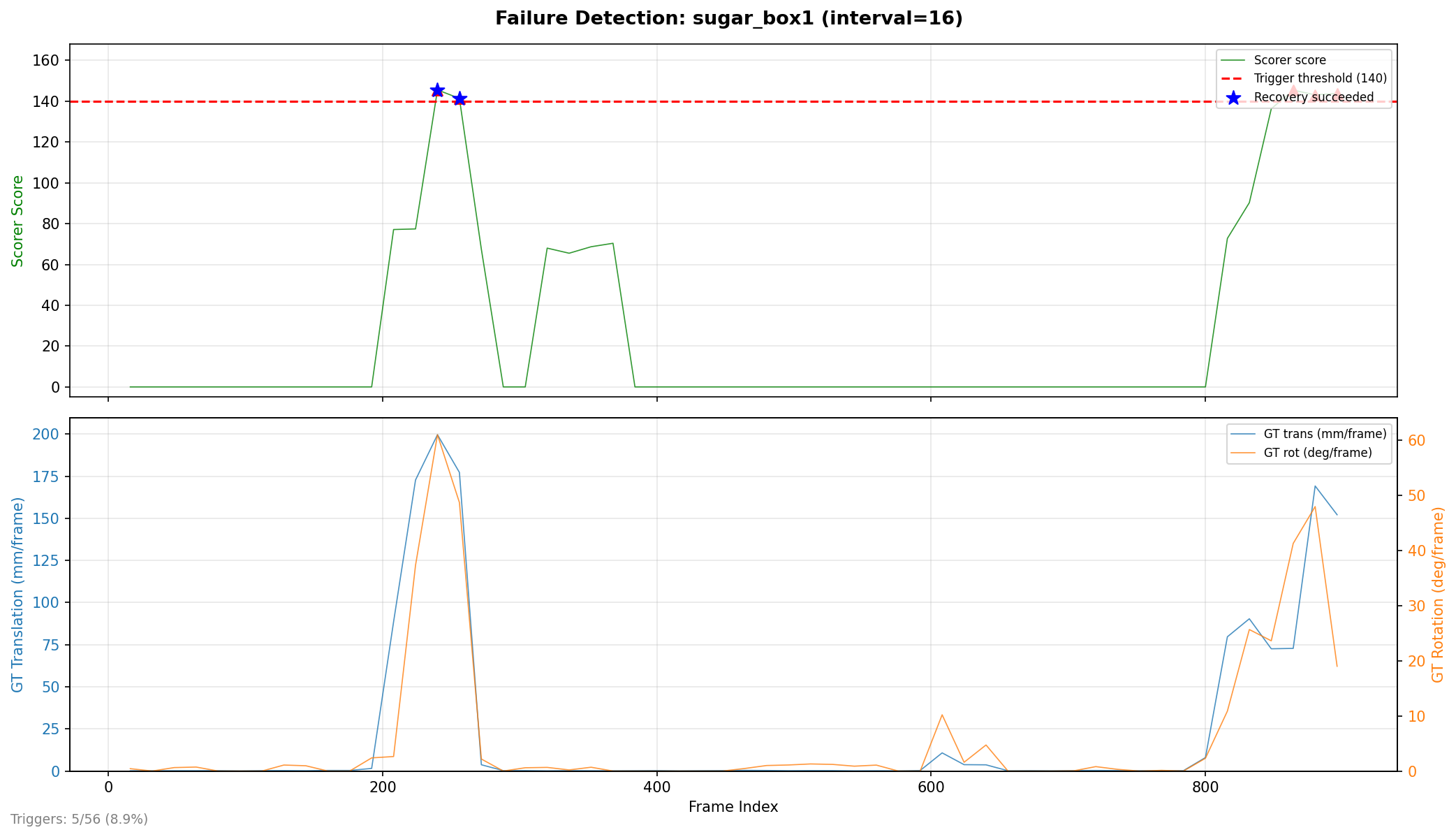}\\
         Case 1: Sugar box tracking \\
         \includegraphics[width=0.75\linewidth]{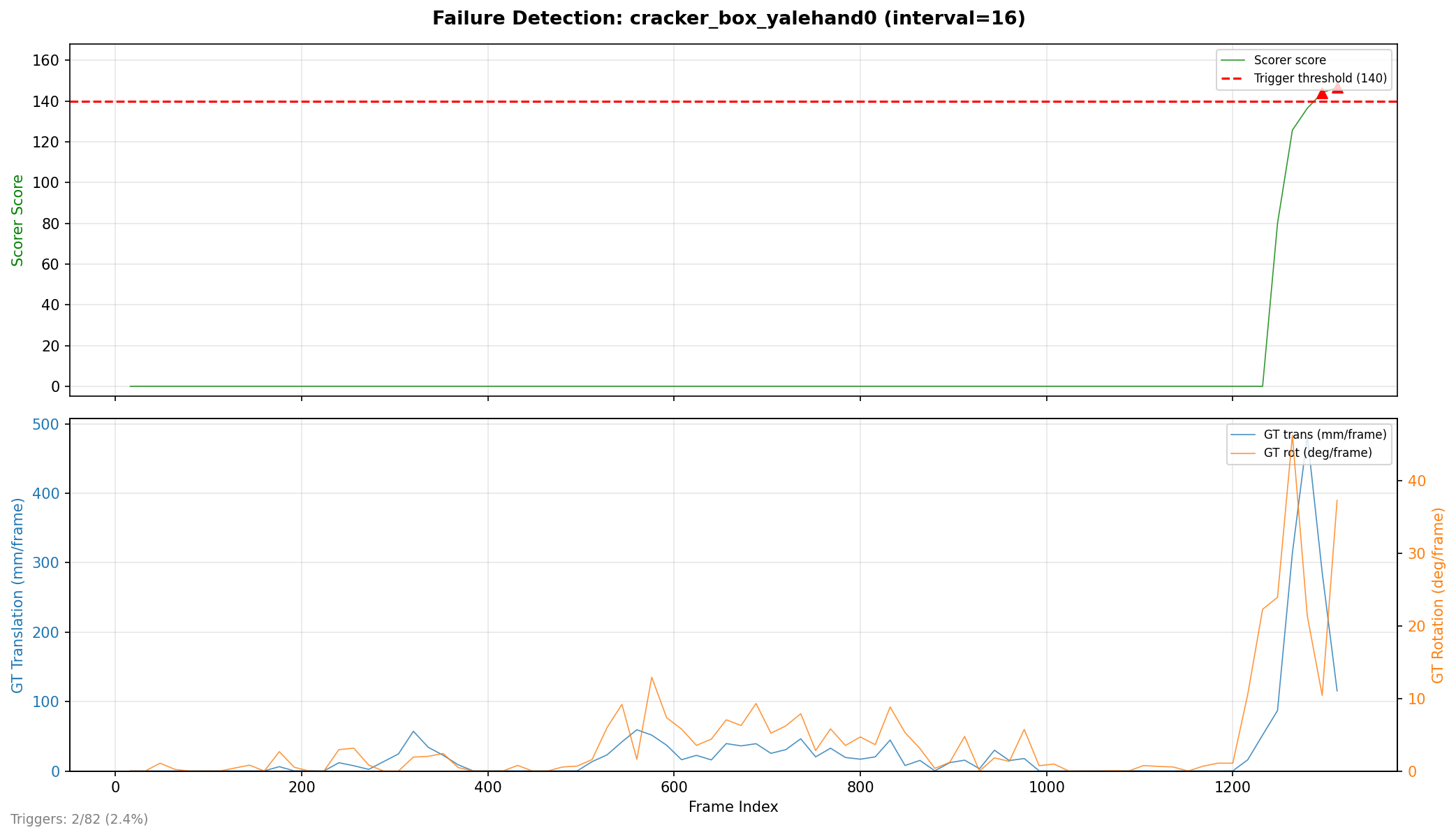}\\
         Case 2: Cracker box tracking\\
    \end{tabular}
    \caption{The relationship between the score function and the translation and rotation changes of the groundtruth. When the objects have rapid motions, it will be reflected in the score functions. We visualized the $\tau_s$ in red dash lines.}
    \label{fig:score-gt-motion}
\end{figure}

We used the score function $\mathcal{F}_S$ of FoundationPose as one condition to trigger the failure recovery. 
We noticed that in general case, the output value of this function $\mathcal{F}_S$ is around $100$.
When there is a large deviation that the pose estimation function $\mathcal{F}_R$ will output an extreme value above $\tau_{s} = 140$.
We emperically used it as a trigger condition.
The second conditions are the differences of the last frame pose $P_{t-1}$ and the estimated pose $P_t$ should be bounded by a range. 
In our experiments, we set the thresholds are $50$ mm and $15$ degree for all testing cases.
Figure~\ref{fig:score-gt-motion} shows the relationship between the output value of the score function $\mathcal{F}_S$ and the translation and rotation changes of the groundtruth.
The threshold $\tau_s$ works on different objects and testing cases.

We followed the 3DGS to use Spherical Harmonics (SH) irradiance model to represent the global lighting parameters.
Recall the Eqn.~\ref{eqn:lighting}, the $u$ represents the pixel in the image domain $\Omega$.
During rendering, it is assigned the base color $I_{CAD}(u)$ from the given CAD model, and normal direction $\mathcal{N}(u, P_{init})$ from the camera view and CAD geometry. 
We use SH coefficients $c$ to compute its radiance field to correct the base color $I_{SH}(u) = L(c, \mathcal{N}(u))$, and $\hat{I}(u, c, P_{init}) = I_{CAD}(u) \cdot I_{SH}(u)$.
In our settings $\textbf{V3}$ and $\textbf{V4}$, we only estimated the lighting coefficients $c$ on the first frame and used $c$ for the rendering on the remaining frames.
We used an AdamW optimizer with the learning rate to $10^{-4}$ and total epochs $200$.
The computation overhead for lighting coefficients is negligible ($< 1$ second).

\section{Experiments on YCB-Video Benchmark}
To further evaluate the robustness and generalization capabilities of our proposed framework, we conducted extensive experiments on the YCB-Video dataset, a widely recognized benchmark for 6DoF object pose estimation and tracking in cluttered real-world scenes.

\begin{figure}[h]
    \centering
    \includegraphics[width=\linewidth]{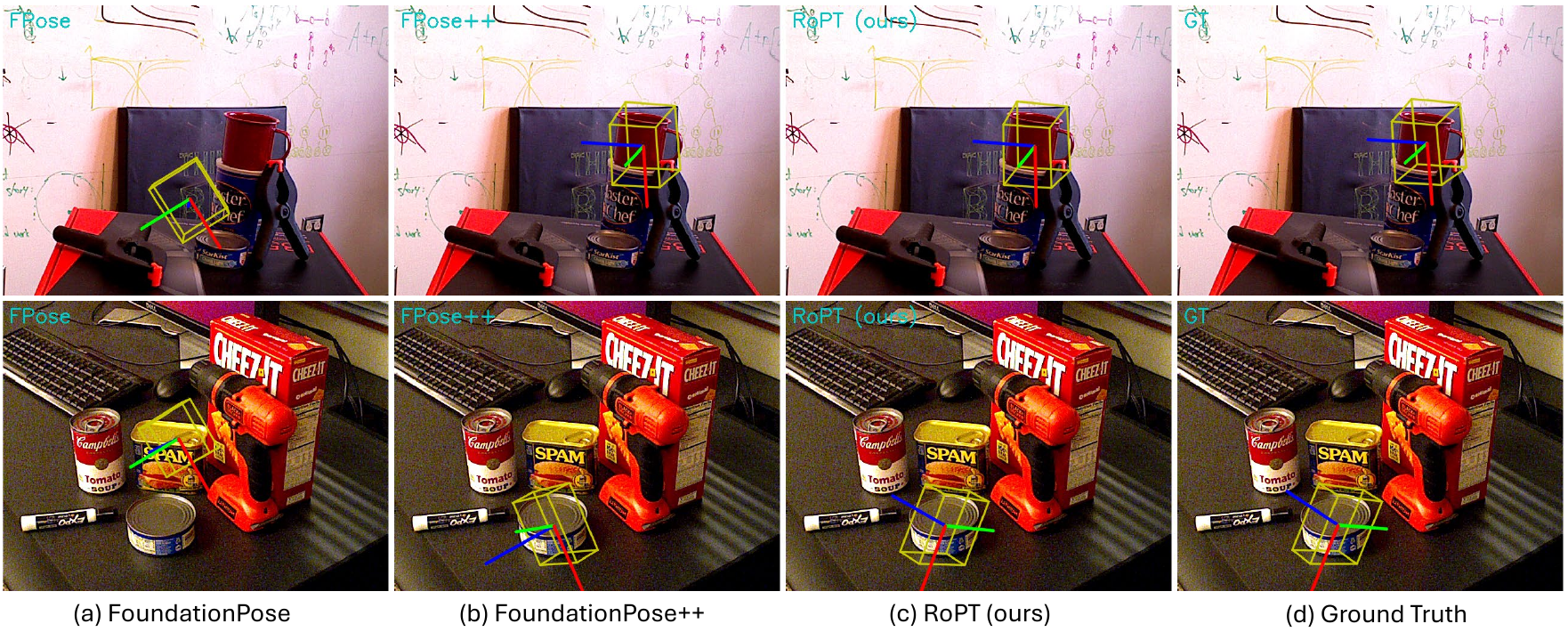}
    \caption{Qualitative comparions on YCB-Video dataset. FoundationPose often lose the object in tracking when camera has a rapid motion.}
    \label{fig:ycbv-qualitative}
\end{figure}

The YCB-Video sequences present significant challenges, including rapid sensor motion, varying lighting conditions, and complex occlusions. 
Our evaluations reveal that FoundationPose often struggles to maintain accurate tracking in these real-world conditions, frequently exhibiting temporal drift or total failure when the object geometry becomes partially obscured. 
In contrast, our method demonstrates superior stability and adherence to the object boundaries despite the lack of depth cues. 
Te quantitative results, detailed in Table~\ref{tab:ycbv-cmp}. We report the average of the three object tracking results in each video. 
Our model outperforms the two state-of-the-art open source code FoundationPose~\cite{wen2024foundationpose} and FoundationPose++~\cite{Wenhao_Yan_and_Jie_Chu_FoundationPose_2025}.
The qualitative results are shown in Figure~\ref{fig:ycbv-qualitative}.

\begin{table}[h]
    \centering
    \resizebox{\textwidth}{!}{%
    \begin{tabular}{c|cc|cc|cc|cc|cc|cc|cc|cc|cc|cc|cc|cc}
    \hline
         \multirow{2}{*}{Video} & \multicolumn{2}{c|}{48} & \multicolumn{2}{c|}{49} & \multicolumn{2}{c|}{50} &\multicolumn{2}{c|}{51} &\multicolumn{2}{c|}{52} &\multicolumn{2}{c|}{53} & \multicolumn{2}{c|}{54} &\multicolumn{2}{c|}{55} &\multicolumn{2}{c|}{56}& \multicolumn{2}{c|}{57} &\multicolumn{2}{c|}{58} &\multicolumn{2}{c}{59}   \\
         & $\mathcal{A}$ & $\mathcal{A}$-S & $\mathcal{A}$ & $\mathcal{A}$-S & $\mathcal{A}$ & $\mathcal{A}$-S & $\mathcal{A}$ & $\mathcal{A}$-S & $\mathcal{A}$ & $\mathcal{A}$-S & $\mathcal{A}$ & $\mathcal{A}$-S & $\mathcal{A}$ & $\mathcal{A}$-S & $\mathcal{A}$ & $\mathcal{A}$-S & $\mathcal{A}$ & $\mathcal{A}$-S & $\mathcal{A}$ & $\mathcal{A}$-S & $\mathcal{A}$ & $\mathcal{A}$-S & $\mathcal{A}$ & $\mathcal{A}$-S\\
    \hline
        FPose & 21.31 & 41.36 & 71.47 & 82.01 & 80.91 & 88.98 & 96.01 & 97.67 & 86.59 & 91.07 & 53.46 & 69.46 & 72.75 & 75.99 & 56.25 & 58.69 & 96.22 & 97.93 & 72.55 & 76.75 & 53.85 & 60.63 & 22.28 & 33.04 \\ 
        
        FPose++ & 77.87 & 96.04 & 75.90 & 96.79 & \cellcolor{cyan!15}94.98 & \cellcolor{cyan!15}97.09 & 83.95 & 92.25 & \cellcolor{cyan!15}96.84 & 98.24 & 73.64 & 97.99 & 86.07 & 97.83 & 94.93 & 97.55 & \cellcolor{cyan!15}96.42 & 97.96 & 79.85 & 97.85 & 97.74 & 98.49 & 78.24 & 97.98\\ 
        
        \textbf{RoPT (ours)} & \cellcolor{cyan!15}92.76 & \cellcolor{cyan!15}96.95 & \cellcolor{cyan!15}95.71 & \cellcolor{cyan!15}97.49 & 89.55 & 91.08 & \cellcolor{cyan!15}97.03 & \cellcolor{cyan!15}98.15 & 96.63 & \cellcolor{cyan!15}98.32 & \cellcolor{cyan!15}78.05 & \cellcolor{cyan!15}98.00 & \cellcolor{cyan!15}96.96 & \cellcolor{cyan!15}98.12 & \cellcolor{cyan!15}96.43 & \cellcolor{cyan!15}98.15 & 96.37 & \cellcolor{cyan!15}98.01 & \cellcolor{cyan!15}91.56 & \cellcolor{cyan!15}97.91 & \cellcolor{cyan!15}97.84 & \cellcolor{cyan!15}98.60 & \cellcolor{cyan!15}96.76 & \cellcolor{cyan!15}98.13 \\ 
       
    \hline
    \end{tabular}
    }
    \caption{Quantitative Comparisons on the YCB-Video dataset.}
    \label{tab:ycbv-cmp}
\end{table}

\begin{table}[h]
    \centering
    \resizebox{0.8\textwidth}{!}{%
    \begin{tabular}{c|cc|cc|cc|cc}
    \hline
         \multirow{2}{*}{Peak Motion Speed} & \multicolumn{2}{c|}{$10 ~cm/s$} & \multicolumn{2}{c|}{$15~cm/s$} & \multicolumn{2}{c|}{$20~cm/s$} &\multicolumn{2}{c}{Average}\\
         & ADD & ADD-S & ADD & ADD-S & ADD & ADD-S & ADD & ADD-S\\
    \hline
        RGBTrack & 9.97 & 23.17 & 5.59 & 15.61 & 11.10 & 31.53 & 8.89 & 23.44  \\
        FPose & 58.88 & 80.67 & 51.97 & 66.89 & 51.10 & 78.96 & 53.98 & 75.51 \\
        FPose++ & 61.01 & 79.39 & 52.65 & 75.93 & 53.98 & 82.12 & 55.88 & 79.14 \\
        \textbf{RoPT (ours)} & \cellcolor{cyan!15}61.26 & \cellcolor{cyan!15}85.36 & \cellcolor{cyan!15}61.78 & \cellcolor{cyan!15}85.77 & \cellcolor{cyan!15}57.37 & \cellcolor{cyan!15}85.85 & \cellcolor{cyan!15}60.13 & \cellcolor{cyan!15}85.66\\
       
    \hline
    \end{tabular}
    }
    \caption{Quantitative Comparisons on the new Pose Tracking dataset.}
    \label{tab:new-pose-tracking-cmp}
\end{table}

\begin{figure}
    \centering
    \resizebox{\textwidth}{!}{%
    \begin{tabular}{cc}
         \includegraphics[width=0.5\linewidth]{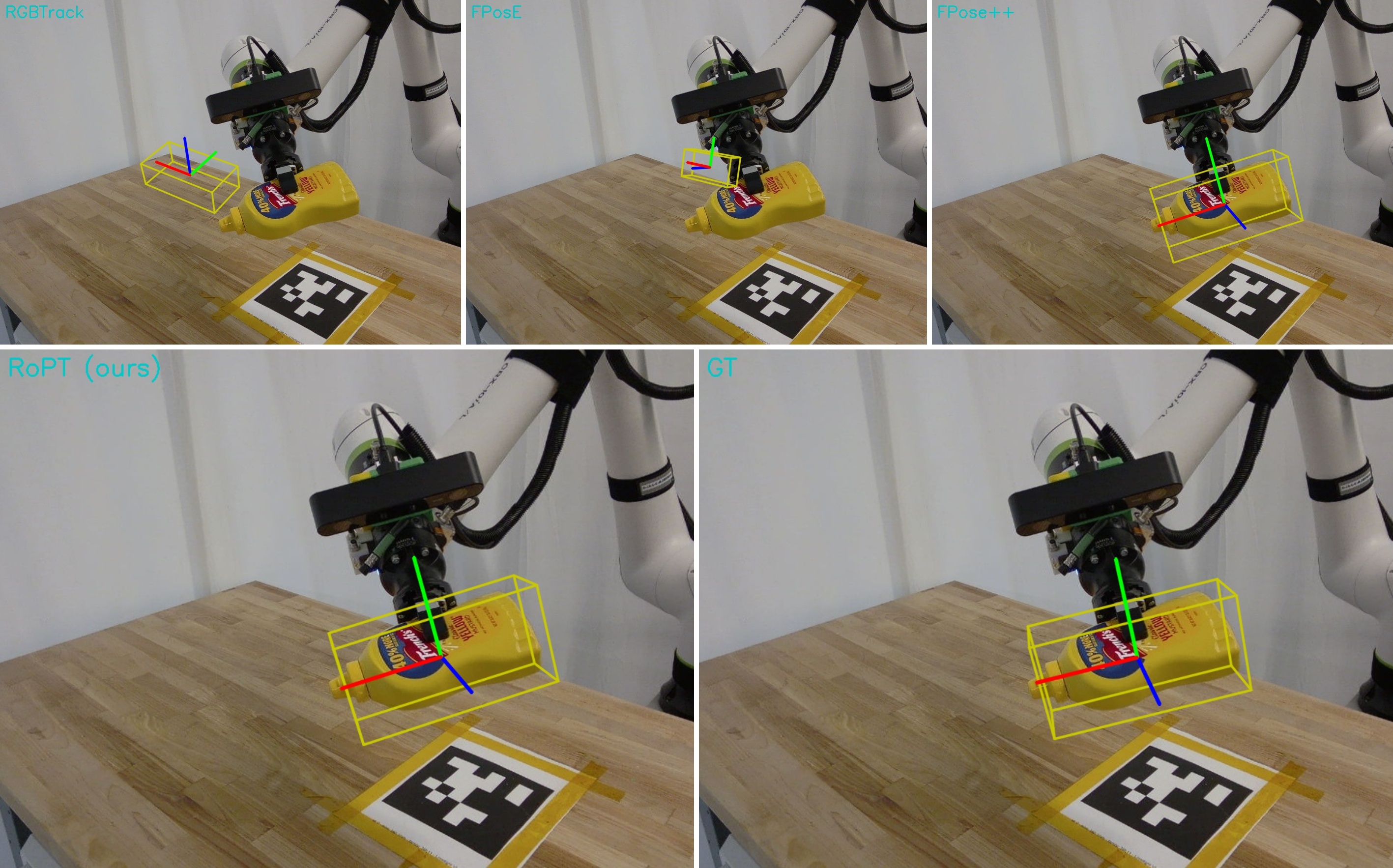}&  
         \includegraphics[width=0.5\linewidth]{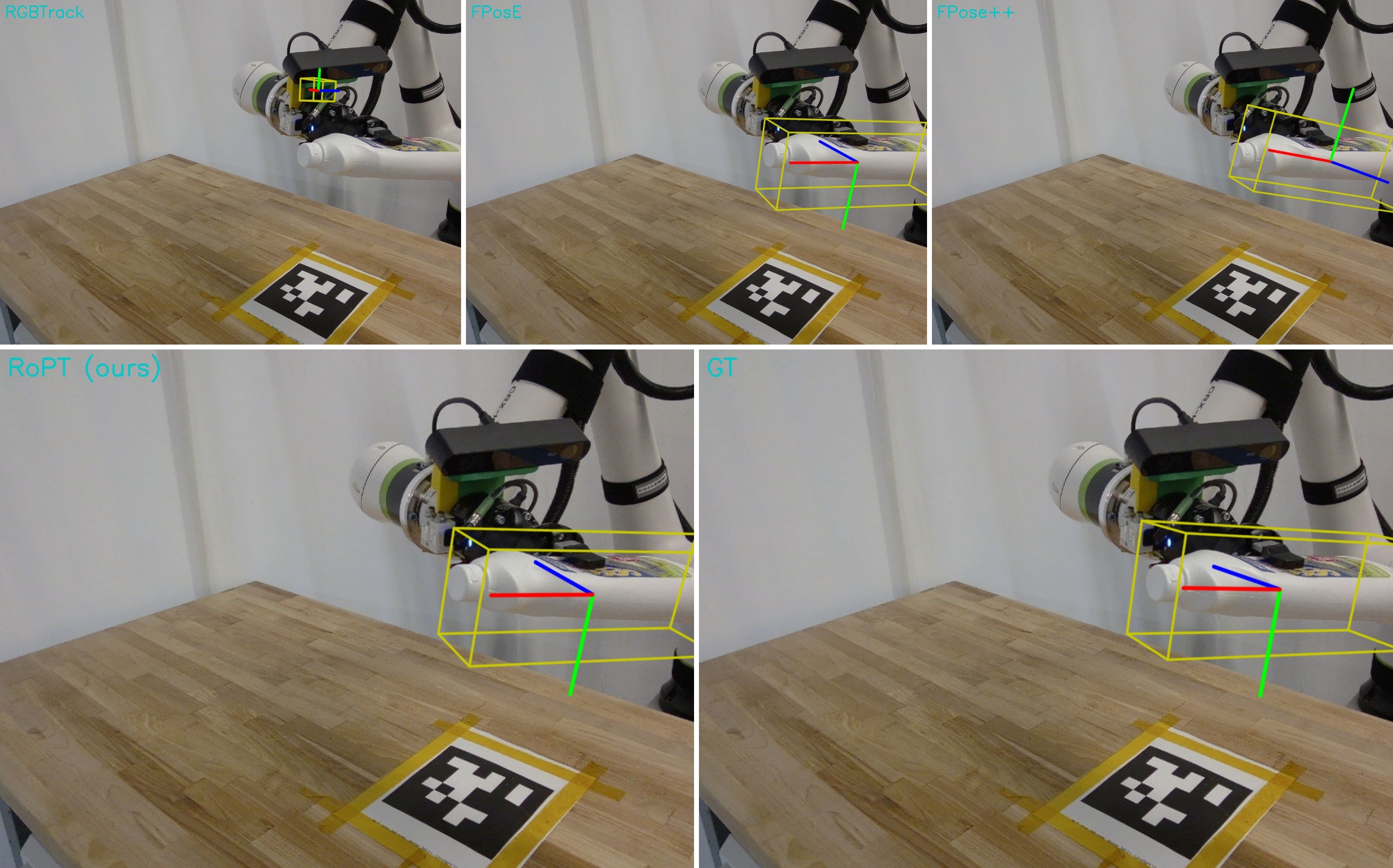}\\
         (a) RGBTrack~\cite{guo2025rgbtrack} and FPose~\cite{wen2024foundationpose} failed on rapid motion. & (b) FPose++~\cite{Wenhao_Yan_and_Jie_Chu_FoundationPose_2025} predicted wrong orientation.
    \end{tabular}
    }
    
    \caption{This figure details qualitative results on the new pose tracking benchmark, where a robotic arm controls the object at real-world speeds. Existing methods frequently struggle to estimate correct translation and orientation.}
    \label{fig:new-pose-tracking}
\end{figure}

\section{Real-World Robotic Demonstrations}

To validate the practical utility of RoPT, we conducted a series of real-world experiments using a hardware with FANUC CRX robot, and an Astra 2 RGB-D sensor in the third-view. While standard pick-and-place tasks are discussed in the main manuscript, we further evaluated and discussed the performance of RoPT in real-world 6-DoF pose tracking and Failure-Aware Self-Recovery.

\subsection{New 6-DoF Pose Tracking Benchmark}
We programmed the robotic arm grasping the target object to execute random trajectories with variable speeds and directions, capping the maximum velocity at 10 cm/s, 15 cm/s, and 20 cm/s. The camera recording frame rate is fixed at 25 fps. We leveraged the calibrated Tool Center Point location of the robotic arm as the 6-DoF pose ground truth to rigorously evaluate both state-of-the-art methods and our proposed approach. We established this pose tracking benchmark to accurately capture the highly dynamic kinematics inherent to real-world robotic operations, addressing a critical limitation of existing benchmarks which predominantly feature artificially slow movements. Ultimately, this benchmark serves as a comprehensive proving ground to validate the performance of pose tracking algorithms, which is an essential step toward demonstrating large-scale data capabilities in complex robotic manipulation and imitation learning tasks.

Table~\ref{tab:new-pose-tracking-cmp} and Figure~\ref{fig:new-pose-tracking} show our benchmark results. The proposed RoPT algorithm overcomes rapid motion challenges, successfully tracking the 6-DoF object pose at both slow and fast speeds.

\subsection{Failure-Aware Self-Recovery Evaluation}
In real-world deployment—such as in imitation learning or Vision-Language-Action (VLA) models—uninterrupted object visibility cannot be guaranteed. Tracking is frequently lost when objects are occluded by environmental obstacles. 
However, a robust 6-DoF tracker must possess two critical capabilities: (1) \textbf{failure awareness}, meaning it recognizes when a reliable 6-DoF pose cannot be computed, thereby preventing erratic outputs that could trigger dangerous robotic actions; and (2) \textbf{self-recovery}, meaning it automatically re-establishes tracking once the target object becomes visible again.
To evaluate these two capabilities, we collected video sequences featuring intentional, dynamic occlusions that block the target object. Existing open-loop estimators are inherently unaware of tracking quality and fail to recover the pose post-occlusion; in contrast, our tracker explicitly detects failures and re-establishes tracking upon object reappearance (see Figure~\ref{fig:recovery}).
Detailed qualitative examples are provided in the supplementary videos.

\begin{figure}
    \centering
    \resizebox{\textwidth}{!}{%
    \begin{tabular}{cc}
         \includegraphics[width=0.5\linewidth]{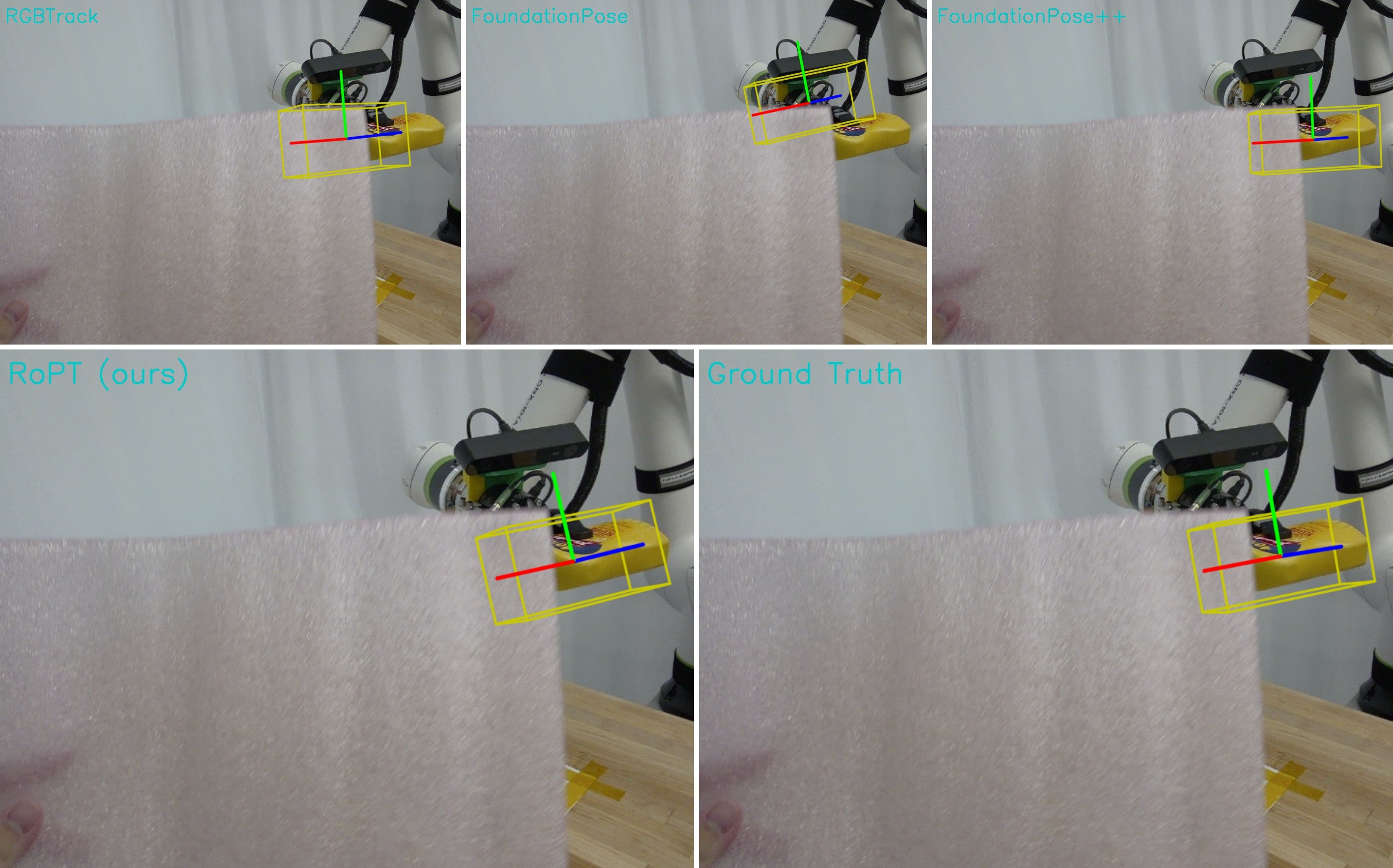}&  
         \includegraphics[width=0.5\linewidth]{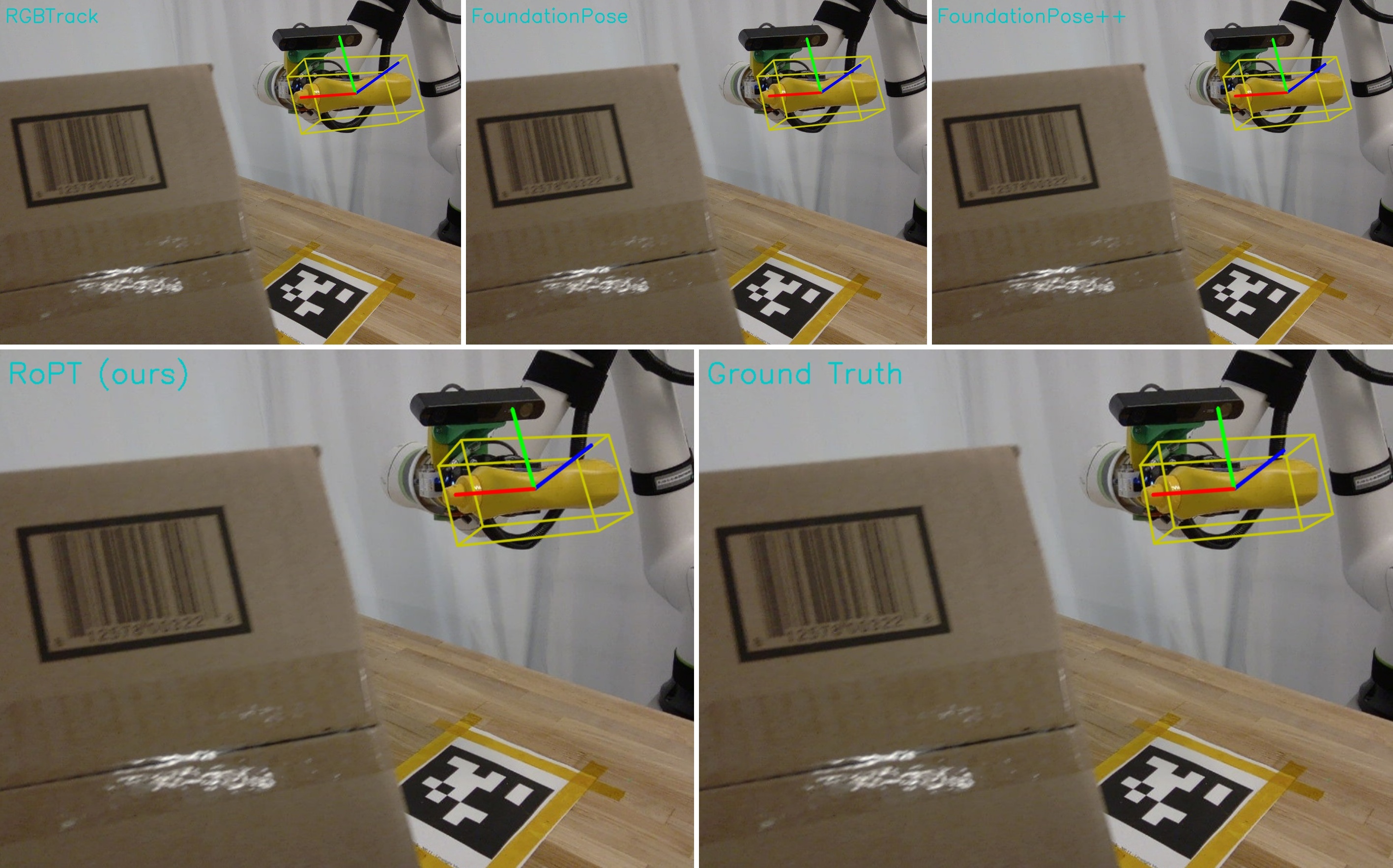}\\
         (a) RGBTrack~\cite{guo2025rgbtrack} and FPose~\cite{wen2024foundationpose} failed on partial occlusion. & (b) An obstacle approaches the object.\\
         \includegraphics[width=0.5\linewidth]{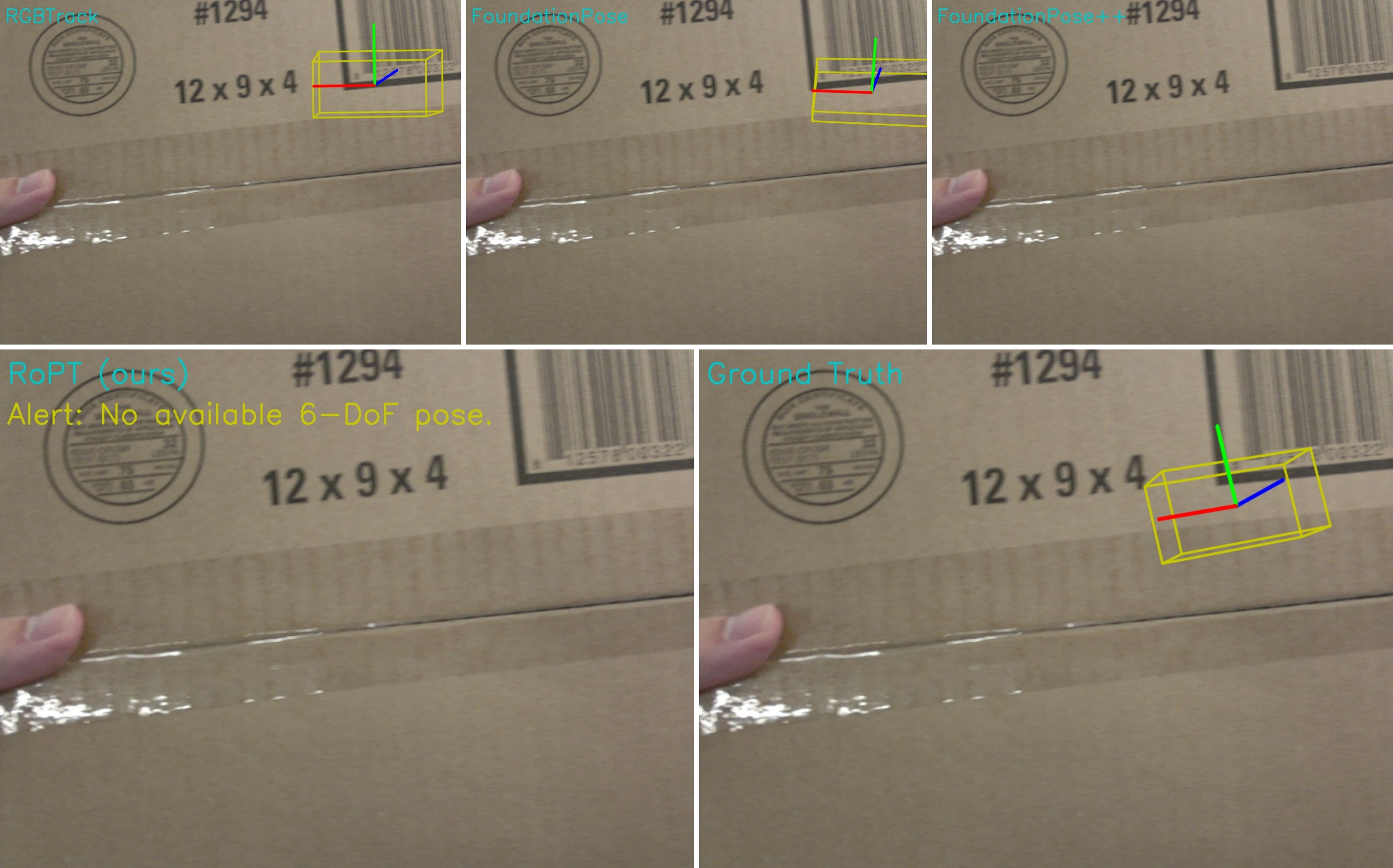} &
         \includegraphics[width=0.5\linewidth]{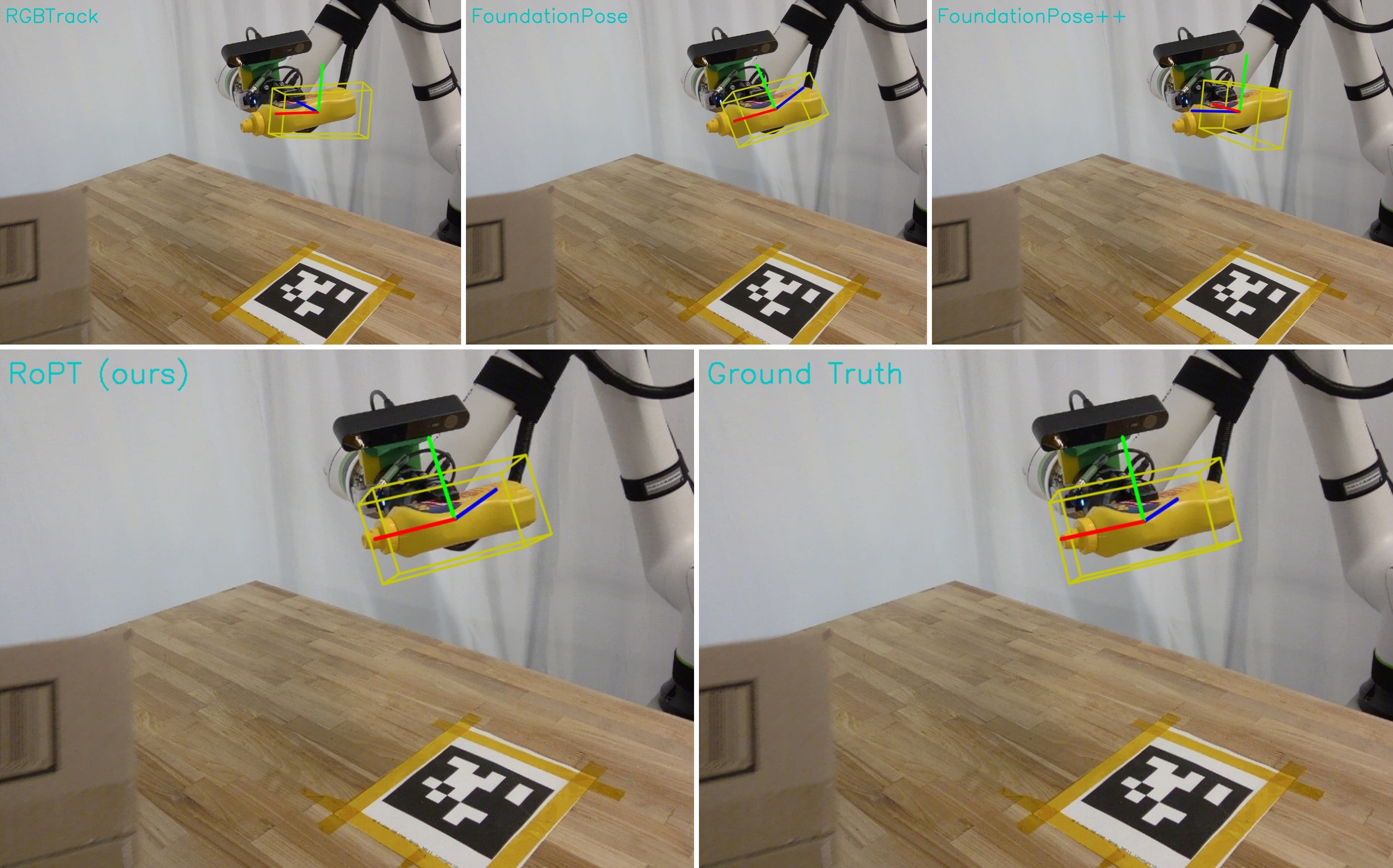}\\
         (c) The obstacle occludes the target object. & (d) Exisiting methods failed to recover 6-DoF pose.
    \end{tabular}
    }
    
    \caption{Failure-Awareness and Pose Recovery Evaluation. Unlike existing open-loop methods that fail to detect unreliable estimates and cannot recover tracking post-occlusion, our method demonstrates both robust failure awareness and self-recovery. (a) Partial occlusion. (b) Approach: An obstacle approaches the target object. (c) Occlusion: The target object is fully occluded by the obstacle. Existing open-loop methods output erratic, unreliable poses, whereas our tracker correctly signals a tracking failure. (d) Pose Recovery: Upon the object's reappearance, our method successfully re-establishes and recovers the accurate 6-DoF pose.}
    \label{fig:recovery}
\end{figure}

\end{appendices}

\end{document}